\documentclass[final,5p,times]{elsarticle}
\usepackage{amsmath,amssymb,amsfonts}
\usepackage{algorithmic}
\usepackage{algorithm}
\usepackage{array}
\usepackage{textcomp}
\usepackage{stfloats}
\usepackage{url}
\usepackage{verbatim}
\usepackage{graphicx}
\usepackage{pgfplots}
\pgfplotsset{compat=1.18}
\usepgfplotslibrary{groupplots}
\usepgfplotslibrary{statistics}
\usepackage{pgfplotstable}
\usepackage{tikz}
\usetikzlibrary{matrix, positioning, calc, backgrounds}
\usepackage{collcell}
\usepackage{hhline}
\usepackage{pgf}
\usepackage{multirow}
\usepackage{booktabs}
\usepackage{colortbl}
\usepackage{pdflscape}
\usepackage{subcaption}
\usepackage{mathptmx}
\usepackage[utf8]{inputenc}
\usepackage[T1]{fontenc}
\usepackage[english]{babel}
\usepackage[hidelinks]{hyperref}
\usepackage{siunitx}
\usepackage{pifont}% http://ctan.org/pkg/pifont
\usepackage{mathrsfs}
\usepackage[percent]{overpic}
\usepackage{xurl}

\usetikzlibrary{arrows.meta, arrows}

\def\colorModel{hsb} %You can use rgb or hsb

\newcommand\ColCell[1]{
  \pgfmathparse{#1<50?1:0}  %Threshold for changing the font color into the cells
    \ifnum\pgfmathresult=0\relax\color{white}\fi
  \pgfmathsetmacro\compA{0}      %Component R or H
  \pgfmathsetmacro\compB{#1/100} %Component G or S
  \pgfmathsetmacro\compC{1}      %Component B or B
  \edef\x{\noexpand\centering\noexpand\cellcolor[\colorModel]{\compA,\compB,\compC}}\x #1
  } 
\newcolumntype{E}{>{\collectcell\ColCell}m{0.4cm}<{\endcollectcell}}  %Cell width

\journal{Measurement}
\myfooter[L]{Author accepted manuscript}
\begin{document}

\begin{frontmatter}
\title{ALICE-LRI: A General Method for Lossless Range Image Generation for Spinning LiDAR Sensors without Calibration Metadata\tnoteref{pubnote}}
\tnotetext[pubnote]{Author accepted manuscript. Published in \textit{Measurement}, volume 294 (2027), article 123662.\newline DOI: \url{https://doi.org/10.1016/j.measurement.2026.123662}.}

\author[citius]{Samuel Soutullo\corref{cor1}}
\ead{s.soutullo@usc.es}
\author[citius,usc]{Miguel Yermo}
\author[usc]{David L. Vilariño}
\author[citius,usc]{Óscar G. Lorenzo}
\author[citius,usc]{José C. Cabaleiro}
\author[citius,usc]{Francisco F. Rivera}

\affiliation[citius]{organization={Centro Singular de Investigación en Tecnoloxías Intelixentes (CiTIUS)},
            addressline={Rúa de Jenaro de la Fuente Domínguez}, 
            city={Santiago de Compostela},
            postcode={15782}, 
            state={A Coruña},
            country={Spain}}

\affiliation[usc]{organization={Departamento de Electrónica e Computación, Universidade de Santiago de Compostela},
            addressline={Rúa Lope Gómez de Marzoa}, 
            city={Santiago de Compostela},
            postcode={15782}, 
            state={A Coruña},
            country={Spain}}

\cortext[cor1]{Corresponding author. Postal address: Centro Singular de Investigación en Tecnoloxías Intelixentes (CiTIUS), Rúa de Jenaro de la Fuente Domínguez, 15782 Santiago de Compostela, A Coruña.}

\begin{abstract}
  3D LiDAR sensors are essential for autonomous navigation, environmental monitoring, and precision mapping, where reliable geometric measurements are required. To efficiently process the massive point clouds generated by these sensors, LiDAR data are often projected into 2D range images that organize points by their angular positions and distances. While these range image representations enable efficient processing, conventional projection methods suffer from fundamental geometric inconsistencies that cause irreversible information loss, compromising high-fidelity applications. We present ALICE-LRI (Automatic LiDAR Intrinsic Calibration Estimation for Lossless Range Images), a metadata-free method that achieves lossless range image generation from calibrated spinning multi-beam LiDAR point clouds without requiring factory calibration files, packet-level data, or beam identifiers. The method is intended for operating regimes in which scanlines contain sufficient returns for reliable parameter estimation. Our algorithm automatically estimates the subset of intrinsic calibration parameters that governs range-image projection in spinning LiDAR measurement systems, including laser beam configuration, angular distributions, and per-beam geometric corrections, enabling metrologically consistent lossless projection and reconstruction with zero point loss. Comprehensive evaluation across the complete KITTI and DurLAR datasets demonstrates that ALICE-LRI achieves perfect point preservation, with zero points lost across all point clouds. Geometric reconstruction deviations remain well below the manufacturer-reported range-performance specifications, establishing geometric losslessness with real-time performance. A compression case study shows improved reconstruction quality and a better quality-compression trade-off in a downstream application. This shift from approximate to lossless LiDAR projections extends range-image-based LiDAR processing to applications requiring high measurement fidelity and preservation of acquired geometric information.
\end{abstract}


\end{frontmatter}

\sloppy

\section{Introduction} \label{sec:introduction}

3D LiDAR sensors have become a key component of modern perception systems, enabling accurate mapping~\cite{haala2008mobile, zhang2014loam}, navigation~\cite{chen2021range, lichtenfeld2024efficient, behley2018efficient}, and scene understanding in autonomous vehicles~\cite{li2020lidar, behley2019semantickitti, milioto2019rangenet++}, as well as in mobile robotics~\cite{weiss2011plant, hutabarat2019lidar}. A widely used class of LiDARs---commonly referred to as rotating or spinning LiDAR sensors---operates by continuously rotating a set of vertically aligned laser beams. As the sensor spins, each beam sweeps out a horizontal arc, producing structured 3D scans of the surrounding environment.

To efficiently process the resulting point clouds, it is common to project them onto 2D range images: dense, grid-like structures where each pixel encodes the distance (range) to the nearest surface along a fixed direction. This projection follows a spherical pattern defined by the sensor geometry, mapping azimuthal angles to horizontal image coordinates and elevation angles to vertical coordinates. Range images are widely used in object detection~\cite{meyer2019lasernet}, semantic segmentation~\cite{wu2018squeezeseg}, upsampling~\cite{shan2020simulation}, and point cloud compression~\cite{zhou2022riddle}, as they enable efficient computation through their regular grid structure and compatibility with established image processing techniques.

This 3D-to-2D projection is often treated as bijective, under the assumption that each point corresponds to a unique combination of range and angular directions. In ideal spinning LiDAR sensors this assumption holds, as the vertical and horizontal angles take discrete values defined by the beam layout and the rotational sampling rate. This enables exact reconstruction of each 3D point from its position and value in the range image. However, this model does not accurately reflect the behavior of real sensors, whose laser beams originate from slightly different positions and orientations due to mechanical constraints. These variations are typically compensated through factory calibration, which adjusts the measured ranges and angles to compute accurate 3D coordinates. Once these corrections are embedded in the Cartesian point cloud, the assumptions of the spherical projection model no longer hold, leading to projection mismatches, pixel collisions, and quantization artifacts when computing range images.

In downstream applications, such artifacts are often ignored or mitigated using ad-hoc techniques. However, in tasks that demand high geometric fidelity even small projection errors may accumulate and degrade performance. A detailed analysis by Wu et al.~\cite{wu2021detailed} demonstrates that standard projection strategies introduce measurable distortions, highlighting the need for more accurate range image generation. While previous work has proposed models that account for laser-specific offsets~\cite{dong2021revisiting}, these approaches rely on manufacturer-supplied lookup tables (LUTs) or packet-level data, limiting their generality. Despite the fundamental importance of this problem for LiDAR-based applications, the challenge of generating truly lossless range images from calibrated spinning LiDAR point clouds without manufacturer metadata has been largely overlooked in the literature.

We introduce ALICE-LRI (Automatic LiDAR Intrinsic Calibration Estimation for Lossless Range Images), a method that estimates the intrinsic calibration parameters governing range-image projection directly from spinning LiDAR point clouds, without requiring factory metadata. These estimated parameters characterize the sensor-specific sampling lattice implicit in calibrated point clouds and are then used to generate range images that are fully consistent with the intrinsic geometry of the sensor.

ALICE-LRI therefore frames range image generation as an inverse sensor-characterization problem. Instead of applying a generic spherical discretization, the method estimates the calibration-dependent projection model needed to preserve the geometric sensing information in the originally acquired point samples during 3D--2D--3D conversion. This avoids projection errors caused by calibration-model mismatch and enables metrologically consistent reconstruction: each reconstructed point maintains a one-to-one link to an originally measured sample instead of an approximate projection bin. The result is a metadata-free, lossless projection algorithm for spinning multi-beam LiDARs, suitable for applications where geometric fidelity is essential. To facilitate adoption, we publish ALICE-LRI as an open-source C++/Python library.\footnote{\url{https://github.com/alice-lri/alice-lri}}

The key contributions of this work are:
\begin{itemize}
    \item The formulation of the fundamental problem of lossless range image generation along with an analysis of how real-world sensor geometry breaks conventional projection methods.
    \item ALICE-LRI, a metadata-free method for lossless range image generation from calibrated spinning multi-beam LiDAR point clouds without requiring factory calibration files, packet-level data, or beam identifiers.
    \item A comprehensive measurement-based evaluation across the complete KITTI and DurLAR datasets, demonstrating zero point loss and geometric reconstruction deviations below relevant manufacturer-reported range-performance specifications, with residuals consistent with input-coordinate precision. We also validate downstream benefits through a compression case study, showing improved reconstruction quality and a better quality-compression trade-off.
\end{itemize}

The remainder of the paper is organized as follows. Section~\ref{sec:related_work} reviews relevant prior work on LiDAR sensor geometry, calibration, metadata-dependent projection, and range-image applications. Section~\ref{sec:the_problem} formulates the lossless range image generation problem and analyzes the gap between ideal and real-world sensor models. Section~\ref{sec:implementation} presents the ALICE-LRI method for automatic parameter estimation and lossless projection. Section~\ref{sec:materials_methods} describes the experimental methodology, datasets, and evaluation metrics. Section~\ref{sec:evaluation} reports comprehensive evaluation results, including parameter estimation accuracy, reconstruction quality, and runtime performance. Section~\ref{sec:application} demonstrates the application and practical benefits of ALICE-LRI through a point cloud compression case study. Finally, Section~\ref{sec:conclusion} concludes the paper, and Section~\ref{sec:future_work} outlines future research directions.

\section{Related Work} \label{sec:related_work}
This section reviews prior work across three fronts. We first discuss LiDAR sensor geometry and calibration. We then review metadata-dependent and approximate range-image projection methods most directly related to ALICE-LRI. We also summarize representative downstream uses of range images, which motivate why projection fidelity matters in practice.

\subsection{LiDAR Sensor Geometry and Calibration} \label{sec:lidar_sensor_geometry_calibration}
Due to manufacturing tolerances and mechanical constraints, real-world LiDAR sensors exhibit geometric non-idealities. These include spatial offsets, angular deviations, and range biases that deviate from theoretical design specifications. LiDAR calibration is a well-established research area concerned with accurately estimating these intrinsic geometric parameters. These parameters are incorporated into the computation of 3D point cloud coordinates, ensuring that output point clouds accurately represent real-world geometry.

A central line of work addresses this forward calibration problem: estimating or refining the physical sensor model itself. Early methods, such as~\cite{muhammad2010calibration}, introduced optimization-based procedures to refine beam orientations and offsets by minimizing point-to-plane distances across multiple scans of structured environments. More extensive models---such as the one proposed in~\cite{glennie2010static} for the Velodyne HDL-64E---incorporated up to six parameters per laser and used planar least-squares adjustments to significantly reduce residual errors and reveal systematic biases across the sensor array. Later, Glennie et al.~\cite{glennie2016calibration} evaluated the factory calibration and long-term stability of the Velodyne VLP-16. They showed that even modest improvements in calibration yield noticeable gains in accuracy, though with limited reusability over time. Other target-based and in-situ methods use pattern planes~\cite{atanacio2011lidar}, automatically detected planar targets~\cite{chen2012onsite}, planar and cylindrical environmental features~\cite{chan2013feature}, local planes, poles, repeated observations, or vehicle motion~\cite{levinson2014unsupervised,chan2015automatic,nouira2016point}.

Recent work continues this forward-calibration direction through optimization-based intrinsic calibration, joint intrinsic/extrinsic calibration, and learning-based extrinsic calibration. Optimization-based intrinsic methods still rely on physical models and geometric constraints to fit calibration parameters~\cite{sun2022intrinsic,wang2024intrinsic}, while joint intrinsic/extrinsic methods use additional sensors and motion observability to calibrate LiDAR-IMU systems~\cite{lv2022observability}. These methods are closely related in the sense that they estimate beam-level or sensor-level geometric quantities, but their objective is metric calibration of the sensor model from richer calibration evidence. Within LiDAR-centered calibration, the recent learning-based literature most directly related to this comparison is mainly concerned with LiDAR-camera extrinsic calibration and broader multi-sensor alignment~\cite{tan2024review}. Such methods estimate relative poses from multimodal data rather than recovering the single-sensor intrinsic projection geometry needed to invert a range-image mapping from calibrated point clouds alone. Consequently, they are not direct experimental baselines for ALICE-LRI: they rely on calibration evidence, motion, or multimodal measurements and optimize metric calibration or inter-sensor alignment, whereas ALICE-LRI estimates the discrete projection geometry implicit in an already calibrated single-sensor point cloud.

Taken together, existing calibration research treats this as a forward problem: estimating parameters from raw data, targets, structured scenes, repeated scans, or multi-sensor motion constraints to improve metric point-cloud accuracy or estimate extrinsic transformations between sensors. However, in many practical scenarios, LiDAR sensors output point cloud coordinates in which the intrinsic calibration has already been applied within the acquisition pipeline. Similarly, most publicly available datasets provide only these final calibrated point clouds, with raw data and beam-level calibration metadata discarded. ALICE-LRI addresses the complementary inverse problem: given a calibrated point cloud, it estimates the subset of sensor-specific intrinsic calibration parameters that determines the projection geometry needed to generate a lossless range image without access to raw packets or calibration metadata. This perspective targets a practical missing-information regime that has received little attention despite its importance for high-fidelity range-image generation.

\subsection{Metadata-Dependent Range-Image Projection} \label{sec:metadata_dependent_projection}
The range-image literature handles sensor-specific projection information through different strategies and under different availability assumptions. Some methods avoid inverse parameter recovery by using information from raw packets, calibration tables, beam identifiers, lookup tables, or preserved scan order. Leveraging such inputs, Dong et al.~\cite{dong2021revisiting} build a ray-wise cylindrical range-image representation from raw spinning-LiDAR scans and use posed range images for registration and reconstruction. When full calibration metadata is unavailable but acquisition order is preserved, scan-unfolding methods can still recover part of the range-view structure. Triess et al.~\cite{triess2020scan} exploit dataset scan ordering to unfold KITTI scans, whereas recent ring-recovery and missing-value filling methods recover ring indices under sequential storage assumptions or rely on laser/ring identifiers when they are available~\cite{chen2025filling}. These approaches are effective under their intended information assumptions, but they do not address the more general setting in which only calibrated Cartesian coordinates are available, without auxiliary metadata or ordering guarantees.

Wu et al.~\cite{wu2021detailed} provide a complementary analysis of information loss in range-image generation. They compare rows defined by laser identifier (PBID) and elevation angle (PBEA), then quantify the reconstruction error obtained after back-projecting the range image to a point cloud. Their results show that the row definition affects quantization error and reconstruction fidelity, especially for sensors with nonuniform beam elevation patterns. Thus, PBID and PBEA are relevant baselines for approximate range-image generation, but they do not provide a lossless point-to-pixel mapping.

ALICE-LRI is positioned between metadata-dependent range-image projection and full LiDAR calibration. Existing work either assumes that the projection geometry is available through metadata or acquisition structure, or estimates a forward calibration model from richer calibration evidence. ALICE-LRI instead targets a narrower but common missing-information regime: inverse intrinsic parameter estimation for geometry-consistent lossless projection when the only available input is a calibrated spinning-LiDAR point cloud.

\subsection{Range Image Applications and Fidelity Requirements} \label{sec:range_image_applications}
The calibration and projection issues discussed above matter because range images remain a common intermediate representation for LiDAR data. The following works illustrate representative settings in which an accurate 3D--2D mapping is useful, and in some cases necessary, for preserving the information content of the scan.

\paragraph{Segmentation}
Segmentation pipelines frequently rely on range images to reduce computation and exploit spatial regularities. Classical methods such as the one proposed by Bogoslavskyi and Stachniss~\cite{bogoslavskyi2017efficient} perform geometric clustering on cylindrical projections. More recent approaches like SqueezeSeg~\cite{wu2018squeezeseg} and RangeNet++~\cite{milioto2019rangenet++} apply convolutional neural networks (CNNs) directly to spherical range images, leveraging their grid structure for efficient inference. RangeNet++, in particular, includes a specialized post-processing module to correct range image discretization errors and smooth out the blurry outputs introduced by CNN inference. Kong et al.~\cite{kong2023rethinking} further identify common artifacts---such as many-to-one conflicts and pixel holes---that arise from discretization and quantization. These efforts highlight the sensitivity of segmentation to projection fidelity and the reliance of current methods on projection-specific corrections or heuristics.

\paragraph{Object Detection}
Range images are widely used for efficient 3D object detection, enabling the use of fast 2D CNNs in place of computationally expensive 3D convolutions. Early work projected LiDAR scans onto cylindrical grids encoding distance and height~\cite{li2016vehicle}, while other methods fused range view with bird's-eye and RGB imagery for enhanced proposal generation~\cite{chen2017multi}. Detection pipelines such as LaserNet~\cite{meyer2019lasernet} and FVNet~\cite{zhou2019fvnet} perform inference directly in the range domain. However, object detection models often overlook the impact of projection fidelity on detection quality, despite facing similar challenges to segmentation. This suggests an opportunity to improve detection robustness by adopting more principled range image generation techniques.

\paragraph{Upsampling}
Range images often also serve as the backbone for LiDAR super-resolution, where low-density scans are upsampled into denser point clouds. Some methods follow a 3D--2D--3D pipeline: sparse scans are projected to 2D, enhanced via neural models, and then backprojected~\cite{shan2020simulation, chen2025srmamba, kwon2022implicit}, despite the 3D--2D--3D pipeline being prone to introducing quantization artifacts. Some recent approaches question this dependency on discrete projections~\cite{tian2022lidar}, but most still assume access to a consistent and accurate mapping between 3D points and 2D pixels.

\paragraph{Odometry and Localization}
Range images also play a central role in LiDAR-based odometry and localization systems, where they serve as structured intermediates for geometric alignment and pose estimation. Behley et al.~\cite{behley2018efficient} proposed a surfel-based SLAM system that leverages dense range images to enable efficient frame-to-model ICP registration. Similarly, Chen et al.~\cite{chen2021range} perform pixel-wise comparisons between live and synthetic range images rendered from a mesh map, enabling GPS-free localization via Monte Carlo inference. More recently, Lichtenfeld et al.~\cite{lichtenfeld2024efficient} introduced a real-time odometry pipeline for dynamic environments that projects both point clouds and residuals onto cylindrical range images. This allows rapid segmentation and tracking of moving objects, which are then filtered out to maintain a clean static map. Across these works, range images consistently boost efficiency and robustness, but their performance hinges on projection quality.

\paragraph{Compression}
A significant family of LiDAR compression methods relies on 2D range images to exploit spatial and temporal redundancies inherent in the structured representation. Projection fidelity is especially important in this class of methods, where information loss directly affects point cloud reconstruction. Early systems~\cite{tu2016compressing} encoded raw packet data into structured range images, enabling efficient compression via JPEG or MPEG. Later methods improved spatial and temporal redundancy exploitation by modeling motion or beam trajectories~\cite{tu2019real, tu2019motion}. Other approaches segment range images into planar patches or predict inter-frame deltas to compress efficiently while preserving geometry~\cite{feng2020real, wang2022efficient}. Recent methods like RIDDLE~\cite{zhou2022riddle} rely on neural predictors and delta encoding in the range domain. Yet all of these methods assume that the input range image is a faithful representation of the 3D scan, a violated condition when projecting from calibrated point clouds using classical methods. As a result, authors either sidestep the issue by operating on raw packet data (when available), or accept minor geometric distortion.

Together, these use cases motivate projection methods that provide accurate and consistent 3D--2D--3D mappings rather than treating range images as approximate representations. ALICE-LRI addresses this need by recovering the sensor-consistent mapping required for lossless conversion from calibrated point clouds, even when raw packets, beam identifiers, and factory metadata are unavailable.

\section{Problem Formulation} \label{sec:the_problem}
This section formulates the core challenge of projecting LiDAR point clouds into 2D range images while preserving all original information. We establish the mathematical foundations by examining an ideal sensor model that enables perfect lossless projection, then analyze how real-world sensor geometry breaks this idealized framework, and finally identify the key parameters that must be recovered to restore losslessness.

\subsection{Range Image Representation} \label{sec:range_image_representation}

Range images provide a structured 2D representation of LiDAR scans by mapping spherical coordinates $(r, \theta, \varphi)$ to a regular grid, where each pixel $(u, v)$ encodes the range $r$ along a discretized angular direction. The projection assumes a spherical sensor model with $L$ laser beams at fixed elevation angles and $H$ uniformly sampled azimuth angles per revolution, creating a mapping between 3D points and image pixels.

\begin{figure}[htpb]
\centering
\includegraphics[width=0.4\textwidth]{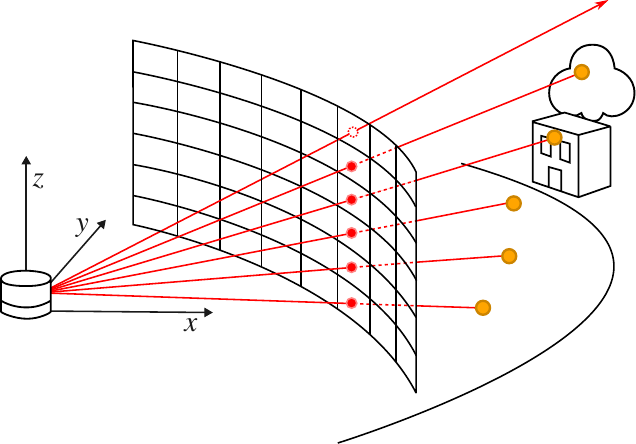}
\caption{LiDAR point cloud projection onto a 2D range image grid based on discretized angular coordinates.}
\label{fig:lidar_to_range_image}
\end{figure}

A lossless projection requires that this mapping is bijective: each pixel corresponds to a unique angular direction, and each measured point maps to exactly one pixel. Figure~\ref{fig:lidar_to_range_image} illustrates this discretization of the sensing sphere into a dense grid structure.

Note that projection fidelity depends critically on how accurately the assumed sensor model reflects the true device geometry. To understand when lossless projection is theoretically possible, we first examine an ideal sensor model that perfectly satisfies the spherical assumption.

\subsection{Ideal Sensor Model} \label{sec:ideal_sensor_model}

Under ideal conditions, a spinning LiDAR sensor exhibits perfect geometric regularity that enables exact lossless projection. We define an ideal sensor as having $L$ laser beams, all originating from a common sensor origin, with each beam oriented at a vertical angle $\varphi^{(l)}$ where $l \in \{0, \dots, L-1\}$. During rotation, the sensor collects $H$ uniformly spaced azimuthal samples per revolution, where each sample $h \in \{0, \dots, H-1\}$ corresponds to an azimuthal angle
\begin{equation}
\theta^{(h)} = \frac{h}{H} \cdot 2\pi.
\end{equation}
For every pair $(l, h)$, the sensor measures a range value $r^{(l,h)}$ corresponding to the distance to the first obstacle encountered in the direction defined by the beam.

Under this model, a point $p_i = (x_i, y_i, z_i)$ captured by the laser beam $l$ at horizontal sample $h$ is uniquely determined from its corresponding triplet $(r^{(l,h)}, \varphi^{(l)}, \theta^{(h)})$, such that
\begin{equation} \label{eq:ideal_coordinates}
\left\{
\begin{aligned}
x_i &= r^{(l,h)} \cdot \cos(\varphi^{(l)}) \cdot \cos(\theta^{(h)}) \\
y_i &= r^{(l,h)} \cdot \cos(\varphi^{(l)}) \cdot \sin(\theta^{(h)}) \\
z_i &= r^{(l,h)} \cdot \sin(\varphi^{(l)}).
\end{aligned}
\right.
\end{equation}
Conversely, given a point $p_i = (x_i, y_i, z_i)$, its spherical coordinates are computed as
\begin{equation} \label{eq:ideal_spherical}
\left\{
\begin{aligned}
r_i &= \sqrt{x_i^2 + y_i^2 + z_i^2} \\
\varphi_i &= \arcsin\left(\frac{z_i}{r_i}\right) \\
\theta_i &= \arctan\left(\frac{y_i}{x_i}\right).
\end{aligned}
\right.
\end{equation}
Since $\varphi_i$ and $\theta_i$ in the ideal model are always drawn from discrete sets of known angles, they must satisfy $\varphi_i \in \{\varphi^{(l)}\}, \forall i$ and $\theta_i \in \{\theta^{(h)}\}, \forall i$. The range $r_i$ can take any real, positive value.

Under these conditions, generating a lossless range image of size $H \times L$ is straightforward. First, compute the spherical coordinates of each point as defined by Equation~\eqref{eq:ideal_spherical}. Then, map each tuple $(r_i, \varphi_i, \theta_i)$ to a pixel $(u_i, v_i)$ in the image using the indices $(l, h)$ of its associated beam angles as
\begin{equation}
    \left\{
    \begin{aligned}
    u_i &\coloneqq h \quad \text{such that} \quad \theta_i = \theta^{(h)} \\
    v_i &\coloneqq l \quad \text{such that} \quad \varphi_i = \varphi^{(l)}.
    \end{aligned}
    \right.
\end{equation}
The corresponding pixel at position $(u_i, v_i)$ is assigned the value $r_i$. Pixels for which no point is present are marked with a sentinel value (e.g., zero or NaN) to indicate absence of measurement.

This ideal model demonstrates that lossless projection is theoretically achievable when sensor geometry exactly matches the spherical assumption. However, real sensors deviate significantly from this idealized framework.

\subsection{Real-World Sensor Model} \label{sec:real_world_model}

Practical LiDAR sensors exhibit geometric non-idealities that fundamentally break the lossless projection capability. Manufacturing tolerances and mechanical constraints cause each laser beam to originate from slightly different physical locations, introducing spatial offsets relative to the sensor origin. Additionally, beams may exhibit unique range deviations due to mechanical and optical factors, and azimuthal offsets from the nominal rotation angle.

These non-idealities render the equations from the ideal model (Equation~\eqref{eq:ideal_coordinates}) insufficient. In response, manufacturers typically define device-specific calibration models, which vary in format but share a common structure: point coordinates depend on beam-specific angles, positional offsets, and range corrections. Velodyne exposes per-laser rotational and vertical corrections together with distance and horizontal/vertical offset corrections~\cite{velodyne_hdl64e_s2_manual}. Ouster computes XYZ coordinates using per-beam altitude and azimuth angles together with a beam-origin offset~\cite{ouster_software_user_manual}. Hesai records per-channel horizontal and vertical angle offsets in an angle-correction file~\cite{hesai_pandar128_manual}. RoboSense exposes corrected vertical and horizontal angle registers for each channel~\cite{robosense_ruby_manual}. Across these models, beam-specific orientation and offset terms are recurring quantities that make the ideal spherical projection insufficient.

The resulting challenge is not that the calibrated Cartesian coordinates are metrically invalid, but that the inverse projection model is incomplete. When these intrinsic projection parameters are not available, standard spherical projection assumes an angular sampling model that no longer matches the measurement geometry that produced the points, introducing projection errors caused by calibration-model mismatch.

\begin{figure*}[htbp] 
    \centering
    \begin{subfigure}[b]{0.48\textwidth}
        \centering
        \includegraphics{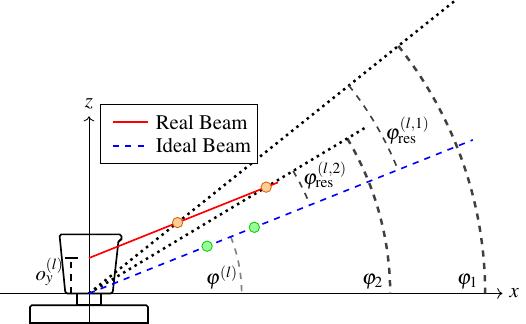}

        \caption{Side view of a LiDAR sensor illustrating two laser beams: an ideal beam (blue, dashed) originating from the sensor's origin, and a real beam (red, solid) with an offset origin. The elevation angles along the ideal beam ($\varphi^{(l)}$) remain constant with respect to the origin of coordinates, while those along the real beam vary ($\varphi_1 \neq \varphi_2$) due to the offset $o_y^{(l)}$ that introduces residual angles $\varphi_{\text{res}}^{(l,h^{(l)})}$ such that $\varphi_1 = \varphi^{(l)} + \varphi_{\text{res}}^{(l,1)}$ and $\varphi_2 = \varphi^{(l)} + \varphi_{\text{res}}^{(l,2)}$.}
    \end{subfigure}
    \hfill
    \begin{subfigure}[b]{0.48\textwidth}
        \centering
        \includegraphics{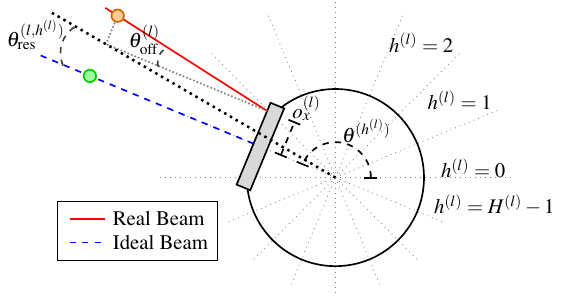}

        \caption{Top view of a LiDAR sensor illustrating two laser beams: an ideal beam (blue, dashed) originating from the sensor's center, and a real beam (red, solid) with a horizontal offset $o_x^{(l)}$ and an azimuthal offset $\theta_{\text{off}}^{(l)}$. The azimuthal angles along the ideal beam remain constant with respect to the origin of coordinates ($\theta^{(h^{(l)})}$), while those along the real beam vary due to the offset that introduces residual angles $\theta_{\text{res}}^{(l,h^{(l)})}$. Further, the azimuthal offset $\theta_{\text{off}}^{(l)}$ shifts the angle of the real beam by a constant amount.}
    \end{subfigure}
    \caption{Illustration of geometric differences between the ideal sensor model (perfectly aligned beams from a common origin) and the real-world model (beams with spatial and angular offsets), highlighting the calibration parameters required for lossless projection.}
    \label{fig:ideal_vs_real_laser}
\end{figure*}

To recover the point-to-pixel mapping, we adopt a projection-oriented formulation that captures the subset of this structure needed for lossless range-image generation. Each laser beam $l$ is characterized by five key parameters, as illustrated in Figure~\ref{fig:ideal_vs_real_laser}: a vertical angle $\varphi^{(l)}$, spatial offsets $(o_x^{(l)}, o_y^{(l)})$, an azimuthal offset $\theta_{\text{off}}^{(l)}$, and a horizontal resolution $H^{(l)}$ representing the number of azimuth samples per revolution for beam $l$. Although many practical sensors employ uniform horizontal resolution across all beams, we allow beam-specific resolutions for generality. Spatial corrections along the forward-backward axis are omitted because they change the measured range $r_i$ but do not introduce the angular residuals that are central to our projection method. Without loss of generality, we assume that beam indices are ordered by increasing vertical angle,
\begin{equation} \label{eq:beam_ordering}
\varphi^{(0)} < \varphi^{(1)} < \cdots < \varphi^{(L-1)}.
\end{equation}

Note that our formulation focuses only on the subset of parameters required to recover a bijective point-to-pixel mapping from calibrated point clouds. This projection-oriented parameterization is deliberately narrower than a complete factory calibration model: it retains the beam-specific orientation and offset terms that affect the inverse mapping, together with $L$ and $H^{(l)}$ as the discrete sampling lattice required to assign points to range-image rows and columns. We intentionally do not model range bias and scale, intensity calibration, temperature drift, encoder-axis misalignment, timing or deskew corrections, boresight/extrinsic parameters, or beam footprint effects. These factors are important for full sensor calibration, but they affect metric calibration rather than the discrete point-to-pixel assignment.

Under this calibrated model, a 3D point $p_i = (x_i, y_i, z_i)$ measured by beam $l$ includes additional residual terms:
\begin{equation} \label{eq:real_coordinates}
    \left\{
    \begin{aligned}
    x_i &= r^{(l,h^{(l)})} \cdot \cos\left( \varphi^{(l)} + \varphi_{\text{res}}^{(l,h^{(l)})} \right) \cdot \cos\left( \theta^{(h^{(l)})} + \theta_{\text{off}}^{(l)} + \theta_{\text{res}}^{(l,h^{(l)})} \right) \\
    y_i &= r^{(l,h^{(l)})} \cdot \cos\left( \varphi^{(l)} + \varphi_{\text{res}}^{(l,h^{(l)})} \right) \cdot \sin\left( \theta^{(h^{(l)})} + \theta_{\text{off}}^{(l)} + \theta_{\text{res}}^{(l,h^{(l)})} \right) \\
    z_i &= r^{(l,h^{(l)})} \cdot \sin\left( \varphi^{(l)} + \varphi_{\text{res}}^{(l,h^{(l)})} \right),
    \end{aligned}
    \right.
\end{equation}
where the residual angles $\varphi_{\text{res}}^{(l,h^{(l)})}$ and $\theta_{\text{res}}^{(l,h^{(l)})}$ are derived from the spatial offsets $(o_x^{(l)}, o_y^{(l)})$ such that
\begin{equation} \label{eq:angular_residuals}
    \left\{
    \begin{aligned}
    \varphi_{\text{res}}^{(l,h^{(l)})} &= \arcsin\left( \frac{o_y^{(l)}}{r^{(l,h^{(l)})}} \right) \\
    \theta_{\text{res}}^{(l,h^{(l)})} &= \arcsin\left( \frac{o_x^{(l)}}{\rho^{(l,h^{(l)})}} \right),
    \end{aligned}
    \right.
\end{equation}
where $\rho^{(l,h^{(l)})}$ is the distance in the XY plane of point $p_i$, measured by beam $l$ at horizontal sample $h^{(l)}$, defined as
\begin{equation}
\rho^{(l,h^{(l)})} = \sqrt{x_i^2 + y_i^2} = r^{(l,h^{(l)})} \cdot \cos\left( \varphi^{(l)} + \varphi_{\text{res}}^{(l,h^{(l)})} \right).
\end{equation}

The azimuthal sampling angles $\theta^{(h^{(l)})}$ are defined similarly to the ideal model, but with beam-specific horizontal resolutions $H^{(l)}$ such that
\begin{equation}
    \theta^{(h^{(l)})} = \frac{h^{(l)}}{H^{(l)}} \cdot 2\pi.
\end{equation}

Because of these additional corrections, using the point coordinates to compute the spherical angles $\varphi_i$ and $\theta_i$ as shown in Equation~\eqref{eq:ideal_spherical} no longer yields results in the discrete sets $\{\varphi^{(l)}\}$ and $\{\theta^{(h^{(l)})}\}$ defined in the ideal model. Instead, they are continuous values defined as:
\begin{equation} \label{eq:real_spherical}
    \left\{
    \begin{aligned}
    \varphi_i &= \varphi^{(l)} + \varphi_{\text{res}}^{(l,h^{(l)})} \\
    \theta_i &= \theta^{(h^{(l)})} + \theta_{\text{off}}^{(l)} + \theta_{\text{res}}^{(l,h^{(l)})}.
    \end{aligned}
    \right.
\end{equation}

As a result, even under the best circumstances, a point will not be projected to the exact center of its corresponding pixel, reducing geometric precision. This misalignment often leads to multiple points being projected onto the same pixel (collisions), or to empty pixels appearing where measurements should exist (holes). These effects fundamentally prevent standard projection strategies from achieving losslessness.

\subsection{Lossless Range Image Generation} \label{sec:lossless_projection}
The prior discussion highlights that the main obstacle to lossless projection is the presence of unknown beam-specific parameters $\{\varphi^{(l)}, o^{(l)}, \theta_{\text{off}}^{(l)}, H^{(l)}\}$. If known, these values can invert the calibration transformations applied to the observed spherical angles, recovering the discrete angular sets required for bijective mapping. In particular, given a point $p_i$ measured by beam $l$ at horizontal sample $h^{(l)}$, it is possible to compute the corrected angles $(\varphi_i', \theta_i')$ using the observed angles $(\varphi_i, \theta_i)$ and the residual terms from Equation~\eqref{eq:angular_residuals} as:
\begin{equation} \label{eq:angle_corrections}
    \left\{
    \begin{aligned}
    \varphi_i' &= \varphi_i - \varphi_{\text{res}}^{(l,h^{(l)})} \\
    \theta_i' &= \theta_i - \theta_{\text{off}}^{(l)} - \theta_{\text{res}}^{(l,h^{(l)})}.
    \end{aligned}
    \right.
\end{equation}
With these corrections, $\varphi_i' \in \{\varphi^{(l)}\}$ and $\theta_i' \in \{\theta^{(h^{(l)})}\}$. This condition guarantees bijective mapping and enables lossless projection. Moreover, the range image dimensions are determined by $L$ (vertical resolution) and the least common multiple of all $H^{(l)}$ (horizontal resolution).

Consequently, recovering these sensor-specific parameters is essential for lossless range image generation. This motivates our approach: we infer the intrinsic projection parameters directly from point cloud data, recovering the sensor-specific mapping needed to avoid calibration-model mismatch during 3D--2D--3D conversion and to generate range images that are intrinsically consistent with sensor geometry.

\section{ALICE-LRI} \label{sec:implementation}
\begin{figure*}[htpb]
    \centering
    \includegraphics[width=0.9\textwidth]{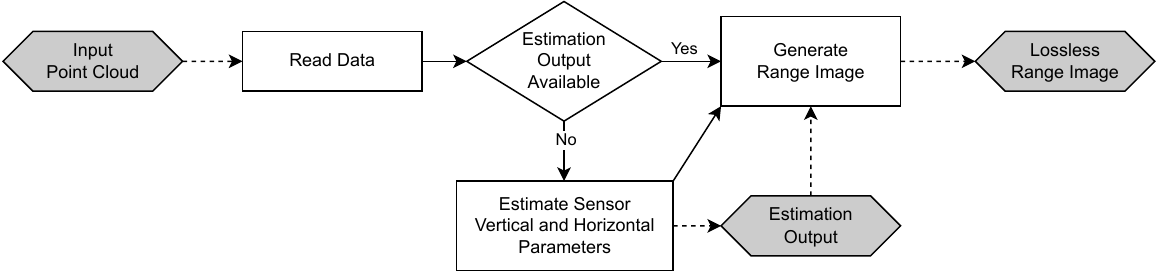}
    \caption{General outline of ALICE-LRI for estimating LiDAR intrinsic parameters and generating lossless range images. Gray diamond-shaped nodes represent data inputs and outputs; rectangular nodes represent processing steps. Solid lines denote execution flow, while dashed lines indicate data flow. The method first estimates sensor parameters from the input point cloud, then uses these parameters to generate range images from subsequent point clouds.}
    \label{fig:general_outline}
\end{figure*}

To solve the problem formulated in Section~\ref{sec:lossless_projection}, we present ALICE-LRI, a general method that automatically recovers the sensor parameters needed for the angle corrections in Equation~\eqref{eq:angle_corrections}. The method operates directly on point cloud data without requiring sensor metadata or scene knowledge. As illustrated in Figure~\ref{fig:general_outline}, ALICE-LRI comprises two main stages: parameter estimation and lossless projection.

In the first stage, the intrinsic parameters are automatically estimated from the point cloud data, including the number of laser beams $L$, the vertical angles $\varphi^{(l)}$, the vertical offsets $o_y^{(l)}$, the horizontal angular resolutions $H^{(l)}$, the horizontal offsets $o_x^{(l)}$, and the azimuthal offsets $\theta_{\text{off}}^{(l)}$. In the second stage, the estimated parameters are used to project the point cloud onto a range image. Since parameter estimation needs to be performed only once per device, the recovered parameters can be subsequently reused to generate range images from multiple point clouds acquired with the same hardware.

The core of the method lies in the parameter estimation stage, which is further subdivided into two phases: vertical parameter estimation and horizontal parameter estimation. The vertical phase determines the number of beams $L$, assigns points to each beam, and simultaneously estimates $\varphi^{(l)}$ and $o_y^{(l)}$. The horizontal phase then leverages these assignments to estimate $H^{(l)}$, $o_x^{(l)}$, and $\theta_{\text{off}}^{(l)}$. The following subsections provide a detailed description of these steps.

\subsection{Vertical Parameter Estimation: $L$, $\varphi^{(l)}$, and $o_y^{(l)}$} \label{sec:vertical_parameter_estimation}
As discussed in Section~\ref{sec:real_world_model}, the vertical angle $\varphi_i$ of each point is affected by a beam-specific vertical offset. Substituting the angular residual $\varphi_{\text{res}}^{(l,h^{(l)})}$ from Equation~\eqref{eq:angular_residuals} into Equation~\eqref{eq:real_spherical} yields
\begin{equation} \label{eq:vertical_fit_model}
\varphi_i = \varphi^{(l)} + \arcsin\!\left(\frac{o_y^{(l)}}{r_i}\right).
\end{equation}
The objective of the vertical parameter estimation is to recover the set of scanlines that best fit the observed data under this model.

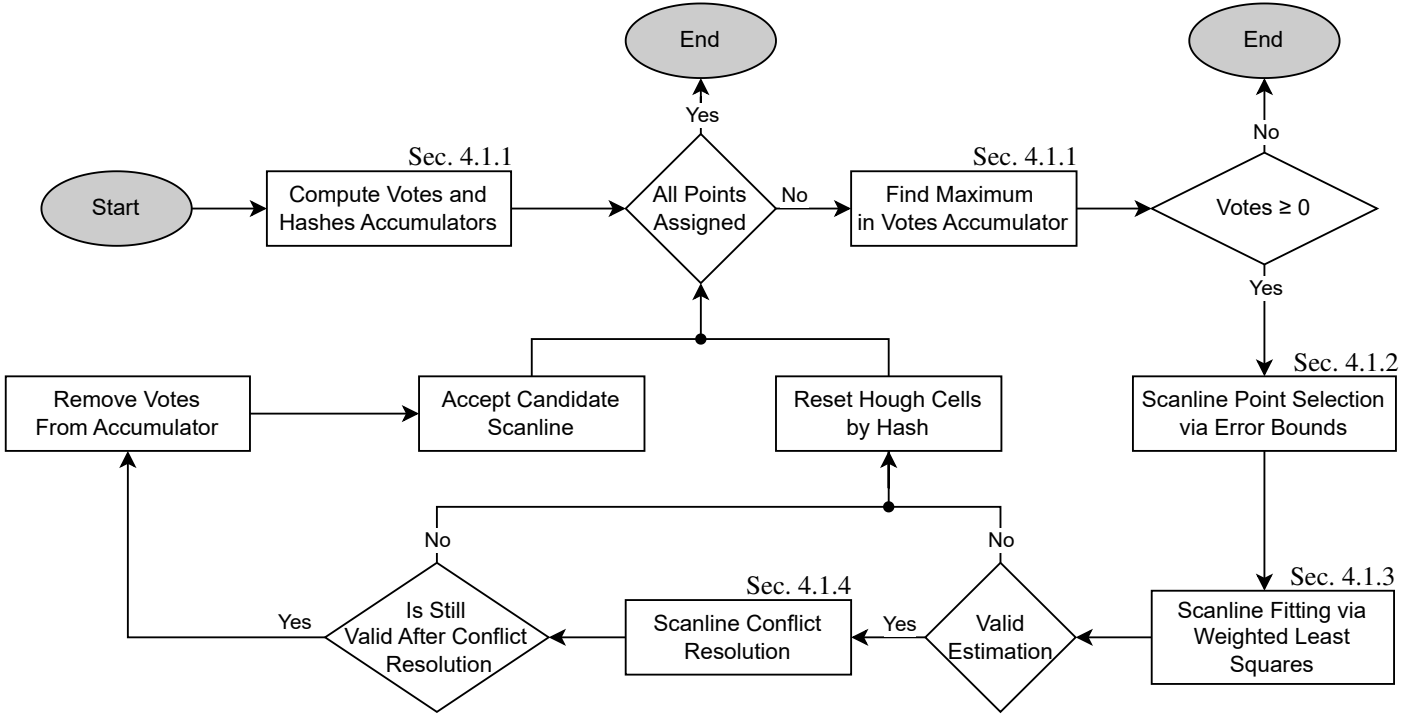
\begin{figure*}[htpb]
    \centering
    \begin{overpic}[width=\textwidth,grid=false]{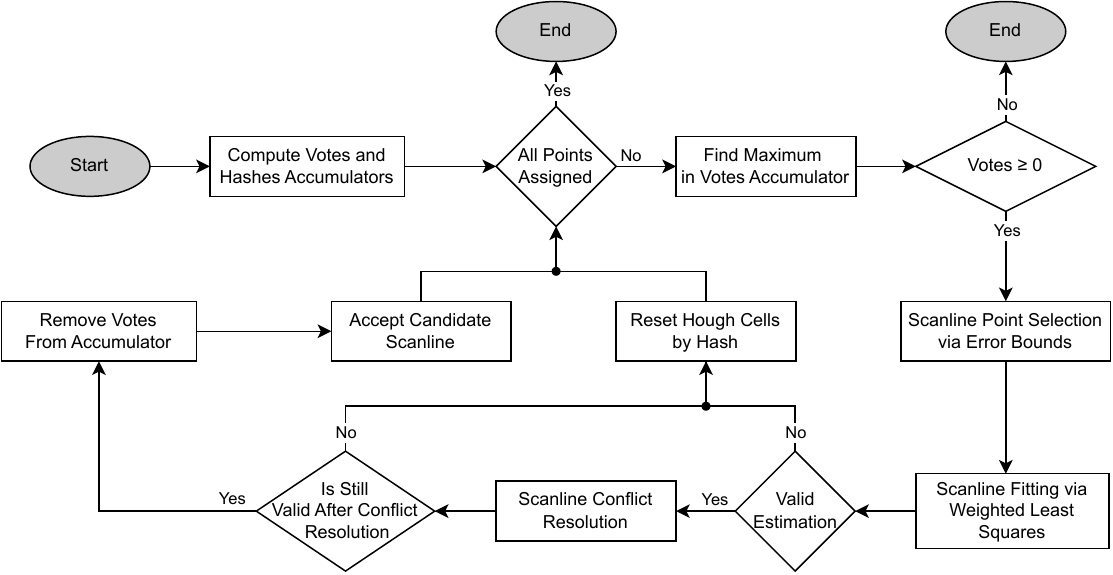}
        \put(29,39.85){Sec.~\ref{sec:hough_transform}}
        \put(69.5,39.85){Sec.~\ref{sec:hough_transform}}
        \put(92.5,25){Sec.~\ref{sec:error_bounds}}
        \put(92.25,9.5){Sec.~\ref{sec:vertical_fitting_wls}}
        \put(53.25,8.9){Sec.~\ref{sec:conflict_resolution}}
    \end{overpic}
    \caption{Flowchart of the vertical parameter estimation algorithm (Sec.~\ref{sec:vertical_parameter_estimation}). The algorithm iteratively identifies candidate scanline parameters using the Hough Transform, selects points consistent with these parameters, fits the scanline to refine parameter values, and resolves conflicts with previously accepted scanlines. The process continues until all points are assigned to scanlines or the search space is exhausted. Key processing steps are annotated with references to their corresponding subsections.}
    \label{fig:vertical_algorithm}
\end{figure*}

To this end, we employ an iterative consensus-based algorithm that progressively identifies candidate scanlines, assigns points to them, and refines their parameters. At each iteration, the algorithm performs four steps: (1) proposes candidate scanline parameters $(\varphi^{(l)}, o_y^{(l)})$, (2) selects points consistent with these parameters, (3) fits the scanline to refine the values of $(\varphi^{(l)}, o_y^{(l)})$, and (4) resolves conflicts with previously accepted scanlines. The process continues until all points are assigned to scanlines or the search space is exhausted. An overview of the proposed algorithm is shown in Figure~\ref{fig:vertical_algorithm}. The subsequent subsections describe each component of the algorithm in detail.

\subsubsection{Candidate Scanline Parameter Identification via the Hough Transform} \label{sec:hough_transform}

The first step in each iteration is to identify candidate scanline parameters $(\varphi^{(l)}, o_y^{(l)})$. To this end, we adopt the Hough Transform~\cite{hough1962method,duda1972use}, a classical feature extraction technique widely used to detect parametric curves in noisy data. While originally developed for line and circle detection in 2D images~\cite{wang1990use,aggarwal2006line}, the method has since been extended to 3D point clouds for extracting structures such as building roofs~\cite{tarsha2007hough}, powerlines~\cite{yermo2024powerline}, and trees~\cite{safaie2021automated}.

The central principle of the Hough Transform is to map each observation in the measurement space to a curve in a discretized parameter space. Observations belonging to the same structure yield curves that intersect at a common point in the parameter space. The coordinates of this intersection define the underlying structure. In our case, each observation $(r_i, \varphi_i)$ is mapped to a curve in the parameter space $(\varphi', o_y)$ obtained by rearranging Equation~\eqref{eq:vertical_fit_model}:
\begin{equation}
\varphi_i' = \varphi_i - \arcsin\left(\frac{o_y}{r_i}\right).
\end{equation}

Thus, points from the same scanline $l$ generate curves that intersect at the true parameters $(\varphi^{(l)}, o_y^{(l)})$. To make these intersections explicit, we discretize the parameter space into an accumulator $\mathbf{A}(\tilde{\varphi}', \tilde{o}_y)$ where
\begin{equation}
\tilde{\varphi}' = \tilde{\varphi}'^{\min} + j \cdot \Delta \tilde{\varphi}', \quad j = 0, 1, 2, \ldots, \left\lfloor \frac{\tilde{\varphi}'^{\max} - \tilde{\varphi}'^{\min}}{\Delta \tilde{\varphi}'} \right\rfloor
\end{equation}
and
\begin{equation}
\tilde{o}_y = \tilde{o}_y^{\min} + k \cdot \Delta \tilde{o}_y, \quad k = 0, 1, 2, \ldots, \left\lfloor \frac{\tilde{o}_y^{\max} - \tilde{o}_y^{\min}}{\Delta \tilde{o}_y} \right\rfloor
\end{equation}
with step sizes $\Delta \tilde{\varphi}'$ and $\Delta \tilde{o}_y$. For each observation $(r_i,\varphi_i)$ and each discretized $\tilde{o}_y$, we compute the corresponding $\tilde{\varphi}'$ and increment the corresponding accumulator cell:
\begin{equation}
\begin{aligned}
&\forall i, \tilde{o}_y: \quad \mathbf{A}(\tilde{\varphi}', \tilde{o}_y) \leftarrow \mathbf{A}(\tilde{\varphi}', \tilde{o}_y) + 1 \\
&\quad \text{where } \tilde{\varphi}' = \left\lfloor \frac{\varphi_i - \arcsin\left(\frac{\tilde{o}_y}{r_i}\right)}{\Delta \tilde{\varphi}'} + 0.5 \right\rfloor \cdot \Delta \tilde{\varphi}'.
\end{aligned}
\end{equation}

In our implementation, the parameter space is discretized with $\tilde{o}_y^{\max} = \min(\min_i r_i,\; O_y)$, $\tilde{o}_y^{\min} = -\tilde{o}_y^{\max}$ and $\Delta \tilde{o}_y = 10^{-3}$ m.
The angular range is fixed to $\tilde{\varphi}' \in [-\Phi,\Phi]$ with $\Delta \tilde{\varphi}' = 10^{-4}$ rad. These values are physically motivated: the vertical offset is bounded by the physical dimensions of LiDAR sensors (we set $O_y=\SI{0.5}{\meter}$, since offsets beyond $\pm O_y$ would be implausible given typical sensor sizes), and it is further clipped by the minimum observed point range to ensure the arcsine argument remains valid. The vertical angles are constrained by $\Phi=\pi/2$, which represents the theoretical maximum range for laser emission. These settings define the initial candidate-generation grid rather than the final parameter precision: accepted candidates are subsequently refined through weighted least squares and checked through uncertainty-based conflict resolution. The Hough discretization therefore balances peak localization and computational tractability, while the continuous fitting stage determines the final estimates.

After all points are processed, peaks in $\mathbf{A}$ correspond to consistent parameter combinations and are selected as candidate scanlines. The iterative phase of the algorithm then begins: at each iteration, the cell with the highest vote count is selected as the next candidate. If accepted, the votes of its contributing points are removed from the accumulator. This strategy, first described as adaptive cluster detection~\cite{risse1989hough}, prevents previously assigned points from influencing later iterations and enables the progressive extraction of further scanlines.

\begin{figure}[h]
    \centering
    \begin{subfigure}[b]{0.23\textwidth}
        \centering
        \resizebox{0.6\linewidth}{!}{\includegraphics{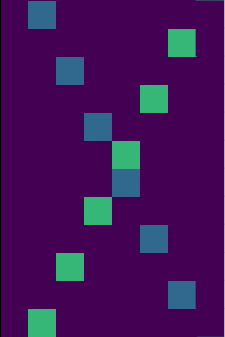}
}
        \caption{Basic Hough Transform strategy}
        \label{fig:discontinuities_regular}
    \end{subfigure}
    \hfill
    \begin{subfigure}[b]{0.23\textwidth}
        \centering
        \resizebox{0.6\linewidth}{!}{\includegraphics{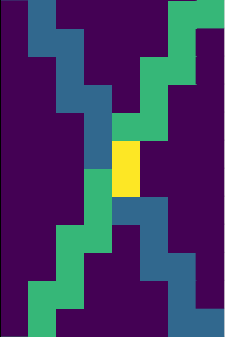}
}
        \caption{Vote-for-discontinuities strategy}
        \label{fig:discontinuities_continuous}
    \end{subfigure}
    \caption{Comparison of basic Hough Transform strategy versus vote-for-discontinuities strategy. The basic strategy (a) misses the intersection peak due to vote fragmentation across cells, while the discontinuities strategy (b) successfully detects it by incrementing intermediate cells when votes skip multiple bins due to steep gradients in the parameter mapping.}
    \label{fig:discontinuities}
\end{figure}

To improve robustness and efficiency, we extend the standard Hough Transform with two modifications. First, we introduce a vote-for-discontinuities strategy: when a vote skips multiple cells in the accumulator (for example, due to steep gradients in the mapping), intermediate cells are also incremented, thereby reducing the risk of fragmented responses (see Figure~\ref{fig:discontinuities}). Second, we maintain a hash accumulator $\mathbf{B}(\tilde{\varphi}', \tilde{o}_y)$ that stores a compact hash of the contributing point indices on each cell. This allows us to quickly determine whether different cells are supported by the same set of points, which is useful for equivalence checks in later stages.

The hashing scheme uses a 64-bit Knuth-style hash~\cite{knuth1973art} applied to each point index and combined by bitwise XOR:
\[
\kappa(i) \coloneqq \big((i+1)\,c\big) \bmod 2^{64},\quad c = 11400714819323198485,
\]
\begin{equation}
\mathbf{B}(\tilde{\varphi}', \tilde{o}_y) \;=\; \bigoplus_{i \in \mathbf{S}(\tilde{\varphi}', \tilde{o}_y)} \kappa(i),
\end{equation}
where $\mathbf{S}(\tilde{\varphi}', \tilde{o}_y)$ is the set of point indices that voted for cell $(\tilde{\varphi}', \tilde{o}_y)$. Cells with identical values therefore share the same contributing set, up to a negligible collision probability.

Despite its utility, the Hough Transform also presents limitations. It identifies candidate parameter locations but does not directly evaluate their quality of fit. Noise may generate spurious peaks or fragment a single scanline into multiple candidates, while sparse scanlines may yield weak responses. Unrelated points can occasionally accumulate votes for an incorrect parameter combination. In addition, achieving higher precision requires finer discretization of the accumulator, which increases both memory and computational costs. For these reasons, we employ the Hough Transform only as the initial stage of our algorithm. The candidate parameters it produces are then refined and validated in subsequent steps.

\subsubsection{Scanline Point Selection via Error Bounds} \label{sec:error_bounds}
Once candidate scanline parameters $(\varphi^{(l)}, o_y^{(l)}) = (\tilde{\varphi}', \tilde{o}_y)$ are identified using the Hough Transform, the next step is to select points that are consistent with these parameters. This filtering stage is essential for the subsequent fitting process, as it ensures that only relevant points contribute to the model.

Due to the finite precision of point cloud coordinates---arising from sensor quantization, rounding, or floating-point representation---points do not generally lie exactly on the theoretical scanline defined by Equation~\eqref{eq:vertical_fit_model}. Some sensors and datasets quantize their coordinates to discrete levels; for example, the KITTI raw dataset~\cite{geiger2013kitti} quantizes coordinates to the nearest millimeter. This coordinate quantization directly propagates as error in the vertical angles $\varphi_i$. To account for this, we derive an angular error bound $\delta\varphi_i$ from the coordinate quantization level. That is, a point $i$ is considered part of scanline $l$ if
\begin{equation} \label{eq:scanline_membership}
\left|\varphi_i - \varphi^{(l)} - \arcsin\!\left(\frac{o_y^{(l)}}{r_i}\right)\right| \leq \delta\varphi_i.
\end{equation}

Since we operate without prior knowledge of the sensor or dataset specifications, we must infer the coordinate quantization level directly from the data itself. We achieve this by examining the minimum non-zero differences between coordinate components across all points in the point cloud.

Assume each Cartesian coordinate is quantized with some unknown step $\Delta_\alpha>0$, i.e.\ $x_i, y_i, z_i \in \Delta_\alpha\mathbb{Z}$. Then every pairwise difference along an axis is an integer multiple of $\Delta_\alpha$, so the minimum nonzero spacing along that axis is at least $\Delta_\alpha$. Let
\[
\Delta^{\min}_x := \min\big\{|x_i-x_j| \neq 0\big\}
\]
be the minimum nonzero difference between the $x$-coordinates (analogously $\Delta^{\min}_y,\Delta^{\min}_z$). We assume that at least one axis exhibits two occupied adjacent quantization levels in the frame\footnote{In practice, where a typical LiDAR frame contains around $10^5$ points, this assumption holds with overwhelming probability, as shown in our experiments (see Section~\ref{sec:evaluation}).}. Under this assumption, we can reliably estimate the quantization level as the smallest of these minimum differences:

\[
\hat{\Delta}_\alpha := \min\{\Delta^{\min}_x,\Delta^{\min}_y,\Delta^{\min}_z\}.
\]
When the observed coordinates are produced by nearest-level quantization, each ideal coordinate can differ from its observed value by at most half a quantization interval. We therefore define the coordinate error bound as $\epsilon = \hat{\Delta}_\alpha/2$. To guard against non-quantized data while covering potential errors introduced by floating-point representation, we use $\epsilon_{\min}=\SI{1e-6}{\meter}$ and clamp $\epsilon \leftarrow \max(\epsilon,\epsilon_{\min})$.

This upper bound on coordinate error $\epsilon$ can then be propagated to derive $\delta\varphi_i$. Let $\mathbf{u_i} = (x_i^*, y_i^*, z_i^*)$ represent the ideal coordinates and $\mathbf{v_i} = (x_i^* \pm \epsilon, y_i^* \pm \epsilon, z_i^* \pm \epsilon)$ the observed coordinates. Defining $\Delta\mathbf{u} = (\pm\epsilon, \pm\epsilon, \pm\epsilon)$, we have $\mathbf{v_i} = \mathbf{u_i} + \Delta\mathbf{u}$. By the triangle inequality,
\[
\|\mathbf{v_i}\| = \|\mathbf{u_i} + \Delta \mathbf{u}\| \leq \|\mathbf{u_i}\| + \|\Delta \mathbf{u}\| = \|\mathbf{u_i}\| + \sqrt{3}\epsilon.
\]
Thus, with $r_i^* = \|\mathbf{u_i}\|$ and $r_i = \|\mathbf{v_i}\|$,
\begin{equation} \label{eq:range_bound}
\delta r = |r_i^* - r_i| \leq \sqrt{3}\epsilon.
\end{equation}
Defining the angular error bound $\delta \varphi_i$ as the difference between the observed and ideal vertical angles, we have:
\[
    \delta \varphi_i = \left|\arcsin\!\left(\frac{z_i^*}{r_i^*}\right) - \arcsin\!\left(\frac{z_i}{r_i}\right)\right|.
\]
Since $\arcsin(z_i/r_i) = \arctan(z_i/\rho_i)$ with $\rho_i = \sqrt{x_i^2+y_i^2}$,
\[
    \delta \varphi_i = \left|\arctan\!\left(\frac{z_i^*}{\rho_i^*}\right) - \arctan\!\left(\frac{z_i}{\rho_i}\right)\right|.
\]
Applying similar reasoning as in Equation~\eqref{eq:range_bound}, we bound
\[
    \delta \rho = |\rho_i^* - \rho_i| \leq \sqrt{2}\epsilon.
\]

By the Mean Value Theorem, for some $\xi$ between $z_i^*/\rho_i^*$ and $z_i/\rho_i$,
\[
    \arctan\!\left(\frac{z_i^*}{\rho_i^*}\right) - \arctan\!\left(\frac{z_i}{\rho_i}\right)
    = \frac{1}{1+\xi^2}\left(\frac{z_i^*}{\rho_i^*} - \frac{z_i}{\rho_i}\right).
\]
Since $\frac{1}{1+\xi^2} \leq 1$, it follows that
\[
    \delta \varphi_i \leq \left|\frac{z_i^*}{\rho_i^*} - \frac{z_i}{\rho_i}\right|
    = \frac{|z_i^{*}\rho_i - z_i\rho_{i}^{*}|}{|\rho_i\rho_{i}^{*}|}.
\]

We now derive an upper bound for $\delta\varphi_i$ by establishing an upper bound on the numerator and a lower bound on the denominator. For the numerator, write
\[
\rho_i^*=\rho_i+\Delta\rho_i,\qquad \Delta\rho_i \in [-\delta\rho,\delta\rho] = [-\sqrt{2}\epsilon,\sqrt{2}\epsilon],
\]
so that
\[
\begin{aligned}
|z_i^*\rho_i - z_i\rho_i^*|
&= \big|z_i^*\rho_i - z_i(\rho_i+\Delta\rho_i)\big| \\
&\hspace{-3em}= \big|(z_i^*-z_i)\rho_i - z_i\,\Delta\rho_i\big| \\
&\hspace{-3em}\le |(z_i^*-z_i)\rho_i| + |z_i \Delta\rho_i| \quad (\text{since }|a - b| \leq |a| + |b|) \\
&\hspace{-3em}\le \epsilon\,\rho_i + |z_i|\sqrt{2}\epsilon\quad  (\text{since }|z_i^*-z_i| \leq \epsilon, |\Delta\rho_i| \leq \sqrt{2}\epsilon) \\
&\hspace{-3em}\le \epsilon\big(\rho_i + \sqrt{2}\,|z_i|\big).
\end{aligned}
\]
For the denominator we have
\[
\rho_i^*=\rho_i+\Delta\rho_i \ge \rho_i-|\Delta\rho_i|\ge \rho_i-\delta\rho,
\]
hence
\[
|\rho_i\,\rho_i^*|=\rho_i\,|\rho_i+\Delta\rho_i|\;\ge\; \rho_i\,(\rho_i-\delta\rho)\;\ge\; \rho_i^2 - \,\sqrt{2}\,\epsilon\rho_i.
\]

Combining the bounds for the numerator and the denominator, we obtain
\begin{equation} \label{eq:phi_bound}
    \delta \varphi_i \leq \frac{\epsilon(\rho_i + \sqrt{2}|z_i|)}{\rho_i^2 - \sqrt{2}\epsilon\rho_i}.
\end{equation}

This bound provides a principled criterion to identify points that belong to a candidate scanline (as defined by Equation~\eqref{eq:scanline_membership}) while explicitly quantifying the uncertainty in $\varphi_i$. Both aspects are crucial for the subsequent fitting stage.

\subsubsection{Scanline Fitting via Weighted Least Squares} \label{sec:vertical_fitting_wls}
\begin{figure*}[htpb]
    \centering
    \includegraphics[width=0.9\textwidth]{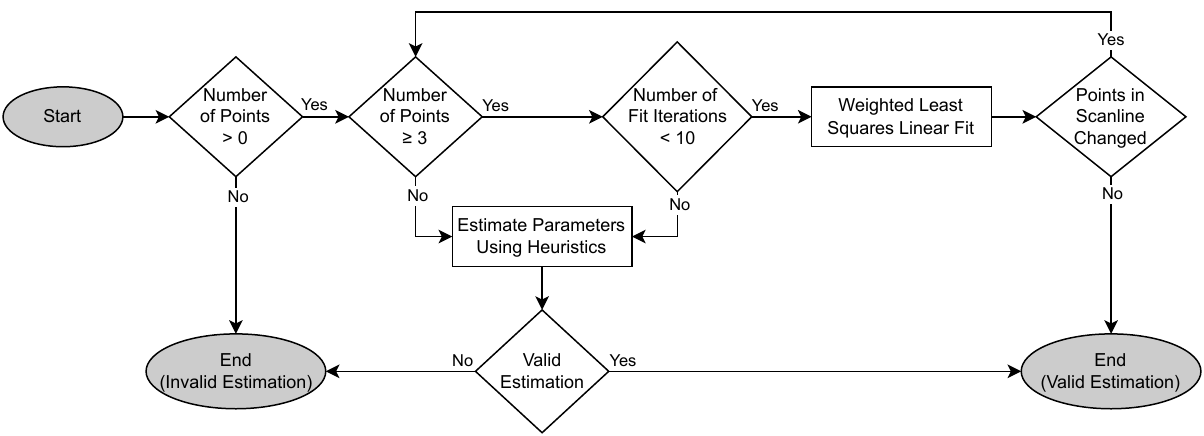}
    \caption{Flowchart of the scanline fitting via weighted least squares procedure (Sec.~\ref{sec:vertical_fitting_wls}). The algorithm performs iterative WLS fitting to refine vertical parameters $(\hat{\varphi}^{(l)}, \hat{o}_y^{(l)})$ and compute point memberships. A heuristic fallback is employed to handle cases where the fitting process fails to converge.}
    \label{fig:vertical_fit}
\end{figure*}

At this stage of the algorithm, we are given a candidate set of points that may form a scanline---i.e., a set of points likely to originate from the same laser beam. The objective is now to estimate the scanline's vertical parameters: the true vertical angle \( \varphi^{(l)} \) and the vertical offset \( o_y^{(l)} \). This is achieved by fitting these points to the model defined in Equation~\eqref{eq:vertical_fit_model}. Figure~\ref{fig:vertical_fit} shows the flowchart of the fitting process.

Note that fitting this model directly is non-trivial as it involves a nonlinear arcsine function, which does not have a closed-form least-squares solution and typically requires iterative optimization techniques. However, we exploit the fact that, in practice, the vertical offset $o_y^{(l)}$ is small relative to the range \( r_i \) of each point (e.g., $o_y^{(l)}$ in centimeters vs. $r_i$ in meters). This allows us to apply the approximation
\begin{equation} \label{eq:arcsin_approx}
    \arcsin(x) \approx x \quad \text{for small } x.
\end{equation}
Applying this approximation to Equation~\eqref{eq:vertical_fit_model}, we obtain
\[
    \varphi_i \approx \varphi^{(l)} + \frac{o_y^{(l)}}{r_i}.
\]
Introducing \( q_i = 1 / r_i \), we rewrite the equation as:
\begin{equation} \label{eq:linear_model}
    \varphi_i \approx \varphi^{(l)} + o_y^{(l)} q_i.
\end{equation}

This linear form enables us to fit the data $(\varphi_i, q_i)$ with a straight line, where the intercept corresponds to $\varphi^{(l)}$ and the slope to $o_y^{(l)}$. However, unlike ordinary least squares, we cannot assume uniform noise across observations: as established in Equation~\eqref{eq:scanline_membership}, each measurement $\varphi_i$ comes with its own error bound $\delta\varphi_i$, leading to heteroscedastic residuals. 

To account for this non-uniform uncertainty, a weighted least squares (WLS) framework is used. We assign each observation a weight $w_i = 1/\delta\varphi_i^2$, which down-weights noisier points relative to more reliable ones. While $\delta\varphi_i$ is not a true standard deviation, Popoviciu's inequality on variances~\cite{popoviciu1935equations} guarantees that for any bounded error $\Delta\varphi_i \in [-\delta\varphi_i, \delta\varphi_i]$, the variance satisfies $\mathrm{Var}(\Delta\varphi_i) \leq \delta\varphi_i^2$. This result justifies treating $\delta\varphi_i^2$ as a conservative upper bound on the residual variance, providing a principled basis for WLS under a heteroscedastic noise model.

Given a set of $n$ pairs $(q_i, \varphi_i)$, the model assumes the linear relation described in Equation~\eqref{eq:linear_model} with the optimal slope $\hat{o}_y^{(l)}$ and intercept $\hat{\varphi}^{(l)}$ obtained by minimizing the weighted residual sum of squares
\[
\min_{\hat{o}_y^{(l)}, \hat{\varphi}^{(l)}} \sum_{i=1}^n w_i R_i^2, \quad \text{where } R_i = \varphi_i - \hat{o}_y^{(l)} q_i - \hat{\varphi}^{(l)}.
\]
This yields closed-form estimators computed using weighted moments:
\[
\hat{o}_y^{(l)} = \frac{S S_{xy} - S_x S_y}{S S_{xx} - S_x^2}, \quad
\hat{\varphi}^{(l)} = \frac{S_{xx} S_y - S_x S_{xy}}{S S_{xx} - S_x^2}, \quad
\]
where $S = \sum w_i$, $S_x = \sum w_i q_i$, $S_y = \sum w_i \varphi_i$, $S_{xx} = \sum w_i q_i^2$, and $S_{xy} = \sum w_i q_i \varphi_i$. To quantify the uncertainty in the estimated parameters, confidence intervals are computed based on the variances of the estimators
\[
\mathrm{Var}(\hat{o}_y^{(l)}) = \sigma^2 \frac{S}{S S_{xx} - S_x^2}, \quad
\mathrm{Var}(\hat{\varphi}^{(l)}) = \sigma^2 \frac{S_{xx}}{S S_{xx} - S_x^2}, \quad
\]
where
\[
\sigma^2 = \frac{1}{n - 2} \sum_{i=1}^n w_i R_i^2
\]
is the weighted residual variance.

We employ the Student's $t$-distribution to construct $95\%$ confidence intervals, using $n-2$ degrees of freedom to reflect the loss of two degrees of freedom from estimating $\hat{o}_y^{(l)}$ and $\hat{\varphi}^{(l)}$. This choice is justified by the small-sample regime, since some scanlines may contain only a few tens of points. Under the assumption of normally distributed residuals within the error bounds, the $t$-distribution is more appropriate than the asymptotic normal approximation.

To evaluate model plausibility, we compute a log-likelihood under the assumption of Gaussian errors with heteroscedastic variance, using the conservative bounds $\delta\varphi_i^2$ as proxies for the variances:  
\[
\log \mathcal{L} = -\frac{1}{2} \sum_{i=1}^n \left[ \log(2\pi \delta\varphi_i^2) + \frac{R_i^2}{\delta\varphi_i^2} \right]
\]

This surrogate log-likelihood captures both the fit quality and the uncertainty structure of the data. Higher values indicate that the observed residuals are small relative to the conservative noise bounds, making the model more statistically plausible. For practical purposes, we define an uncertainty score as the negative log-likelihood,  
\begin{equation} \label{eq:uncertainty_score}
\mathcal{U} = -\log \mathcal{L},
\end{equation}
which enables direct comparison between alternative fits: lower scores indicate better alignment with the observed data and its bounded-error structure.

A single WLS fit would be sufficient if the initial estimates from the Hough transform were already accurate and the corresponding candidate points were correctly assigned. In practice, however, this is not always the case, so a more robust iterative refinement is required. After the initial WLS fit, the estimated parameters $(\hat{o}_y^{(l)}, \hat{\varphi}^{(l)})$ are used to reassess point membership based on the criterion in Equation~\eqref{eq:scanline_membership}, yielding an updated point set. The model is then re-fitted to this refined set, with the process repeated until point assignments stabilize or a maximum of $N_{\mathrm{WLS}}=10$ iterations is reached. This iterative scheme ensures convergence to a stable, self-consistent solution both for parameters and point memberships.

\paragraph{Fallback: Heuristic Estimation}

In some edge cases, the WLS fitting procedure may fail to converge. Namely when the number of candidate points is insufficient or when the convergence is not achieved within the iteration limit. As a last resort for these rare scenarios where the mathematical fitting is insufficient, we introduce a heuristic fallback that leverages assumptions based on the physical arrangement of the laser beams of real LiDAR sensors. Specifically, it assumes that the vertical offset of the laser beam can be approximated as the average offset of the two neighboring lasers, that is $\hat{o}_y^{(l)} = \frac{1}{2} \left( \hat{o}_y^{(l-1)} + \hat{o}_y^{(l+1)} \right)$. This is reasonable, as physically adjacent lasers tend to have similar vertical angles. This method is only triggered when some reliable scanlines have already been computed, as it depends on existing parameter estimates. Once the offset \( \hat{o}_y^{(l)} \) is heuristically determined, the vertical angle \( \hat{\varphi}^{(l)} \) is directly obtained by rearranging Equation~\eqref{eq:linear_model} and computing the mean value.

\begin{equation}
\hat{\varphi}^{(l)} = \frac{1}{N} \sum_{i=1}^{N} \left( \varphi_i - \hat{o}_y^{(l)} q_i \right)
\end{equation}

After either the WLS or the heuristic fit, the algorithm recomputes which points fall within the estimated scanline bounds. If the point set is empty the scanline candidate is marked as invalid and rejected. To avoid redundant processing of identical scanlines detected in multiple Hough cells, once a scanline is rejected we employ the hash accumulator of Section~\ref{sec:hough_transform}, resetting all cells with identical hashes to zero:
\begin{equation} \label{eq:hash_reset}
\forall (u, v) : \mathbf{B}(u, v) = \mathbf{B}(\tilde{\varphi}', \tilde{o}_y) \implies \mathbf{A}(u, v) = 0,
\end{equation}
thereby ensuring that identical point sets are not repeatedly re-evaluated. Conversely, if either the WLS or the heuristic fit is successful, the scanline is passed to the conflict resolution step.

\subsubsection{Scanline Conflict Resolution} \label{sec:conflict_resolution}
\begin{figure*}[htpb]
    \centering
    \includegraphics[width=0.9\textwidth]{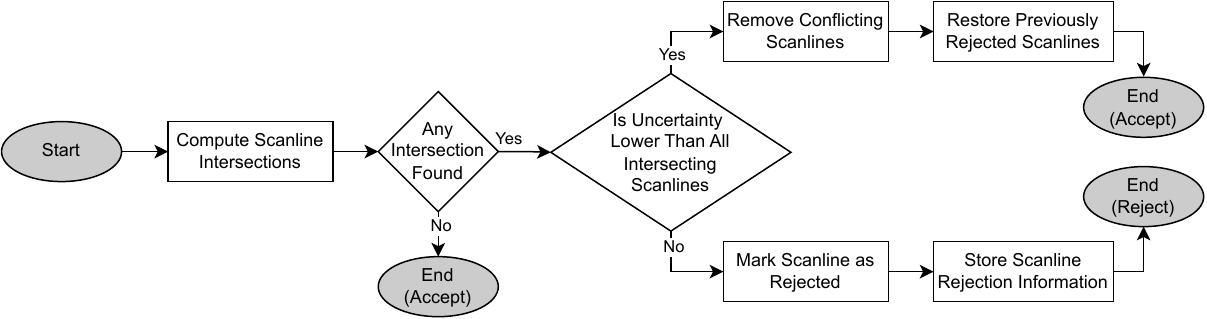}
    \caption{Flowchart of the scanline conflict resolution process (Sec.~\ref{sec:conflict_resolution}). The algorithm checks for intersections between the current scanline candidate and previously accepted scanlines. If intersections exist, uncertainty scores are compared to determine which scanlines to accept or reject. Rejected scanlines are stored for potential recovery if their conflicting counterparts are later invalidated.}
    \label{fig:scanline_conflict}
\end{figure*}

Up to this point, the algorithm has focused on estimating individual scanlines in isolation. Accepting each candidate greedily would risk inconsistencies, such as multiple scanlines claiming the same points or geometrically implausible intersections. These issues are especially likely in later iterations, when candidate scanlines are supported by fewer points and thus become more sensitive to noise or imperfect fitting. To address this, we introduce a conflict resolution step that enforces global consistency and supports backtracking, enabling previously accepted scanlines to be discarded in favor of better alternatives. Its objectives are twofold: (1) to resolve conflicts between the current scanline and those already accepted, thereby preventing inconsistencies in the final set, and (2) to allow recovery from suboptimal local decisions, increasing the likelihood of a globally consistent solution. An overview of the conflict resolution process is shown in Figure~\ref{fig:scanline_conflict}.

Conflicts may arise as point intersections, where at least one point is simultaneously claimed by multiple scanlines, or as geometric intersections, where fitted scanlines (including their confidence intervals) overlap within the valid range of the point cloud. Both checks are performed. If no intersections are found, the scanline is accepted: its parameters are stored, points are assigned, and their votes removed from the Hough accumulator (Section~\ref{sec:hough_transform}); otherwise, conflicts must be resolved.

The conflict resolution process relies on the uncertainty scores computed during fitting, as given by Equation~\eqref{eq:uncertainty_score}. A new scanline with higher uncertainty than any conflicting counterpart is deemed less reliable and rejected. Conversely, if it is more reliable than all conflicting counterparts, it is accepted, and the conflicting scanlines are invalidated. Rejected candidates are not permanently discarded: their parameters and rejection metadata---including the identities of the scanlines causing invalidation---are stored for potential recovery. After invalidating conflicting scanlines, the algorithm iterates over previously rejected candidates. If any no longer exhibit conflicts, they are reintroduced into the Hough accumulator for potential future selection. This backtracking mechanism allows the algorithm to recover from earlier suboptimal choices and promotes convergence to a globally consistent solution.

Once conflict resolution is complete, the new scanline is either accepted or rejected, and the algorithm continues accordingly. When scanlines are rejected, the same hash-based cell reset procedure described in Sec.~\ref{sec:vertical_fitting_wls} is applied to avoid redundant processing of identical scanlines (see Equation~\eqref{eq:hash_reset}).

\subsection{Horizontal Parameter Estimation: $H^{(l)}$, $o_x^{(l)}$, and $\theta_{\text{off}}^{(l)}$} \label{sec:horizontal_estimation}
After estimating the vertical parameters, the algorithm has identified the number of laser beams $L$, the vertical angles $\varphi^{(l)}$, the vertical offsets $o_y^{(l)}$, and the assignments of each point in the cloud to its originating laser beam. The next stage is to estimate the horizontal parameters: the horizontal angular resolution $H^{(l)}$, the horizontal offset $o_x^{(l)}$, and the azimuthal offset $\theta_{\text{off}}^{(l)}$ of each beam. Expanding the second expression in Equation~\eqref{eq:real_spherical}, the observed azimuth $\theta_i$ can be expressed as
\begin{equation}
\theta_i = \theta^{(h^{(l)})} + \theta_{\text{off}}^{(l)} + \theta_{\text{res}}^{(l,h^{(l)})}
         = \frac{h^{(l)}}{H^{(l)}} \cdot 2\pi + \theta_{\text{off}}^{(l)} + \arcsin\!\left(\frac{o_x^{(l)}}{\rho_i}\right).
\end{equation}

\begin{figure*}[h]
\centering
\includegraphics[width=0.9\textwidth]{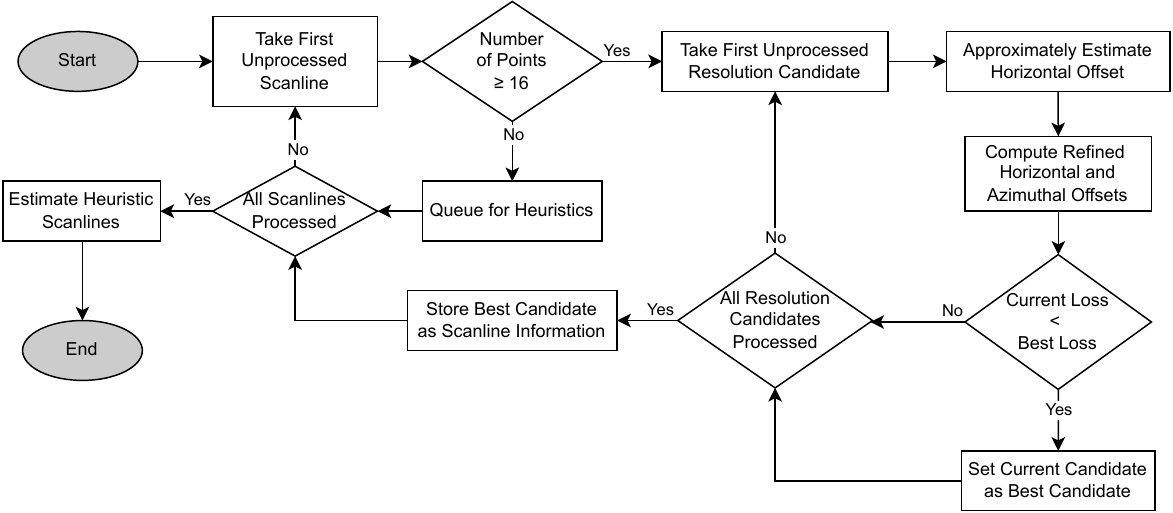}
\caption{Flowchart of the horizontal parameter estimation algorithm (Sec.~\ref{sec:horizontal_estimation}). The algorithm performs an exhaustive search over candidate horizontal resolutions $\hat{H}^{(l)}$ for each scanline, computing horizontal offset $\hat{o}_x^{(l)}$ and azimuthal offset $\hat{\theta}_{\text{off}}^{(l)}$, and selects the parameters that minimize the loss function. For scanlines with insufficient points, a heuristic fallback reuses parameters from previously processed beams.}
\label{fig:horizontal_algorithm}
\end{figure*}

The objective of this step is to recover the set $\{H^{(l)}, o_x^{(l)}, \theta_{\text{off}}^{(l)}\}$ for each beam $l$ such that the scanline observations are best fitted. We treat each scanline independently, without assuming correlations across beams, which enhances robustness to variations in sensor design. An overview of the algorithm is shown in Figure~\ref{fig:horizontal_algorithm}.

Using Equation~\eqref{eq:arcsin_approx}, a simplification is obtained by exploiting that $o_x^{(l)} \ll \rho_i$. Introducing $\omega_i = 1/\rho_i$, we approximate
\begin{equation} \label{eq:theta_approx}
\theta_i \approx \frac{h^{(l)}}{H^{(l)}} \cdot 2\pi + \theta_{\text{off}}^{(l)} + o_x^{(l)} \cdot \omega_i.
\end{equation}

Since $H^{(l)}$ is an integer, with a minimum value equal to the number of points in scanline $l$ and a practical maximum on the order of $10^3$ for LiDAR sensors~\cite{velodyne_hdl64e_s2_datasheet,ouster_os1_datasheet,hesai_pandar128_manual,robosense_ruby_manual}, it is feasible to perform an exhaustive search over candidate resolutions $\hat{H}^{(l)} \in [n^{(l)},H_{\max}]$. For each candidate, we compute the corresponding values of $\hat{o}_x^{(l)}$, $\hat{\theta}_{\text{off}}^{(l)}$, and a loss metric $L(\hat{H}^{(l)})$. The parameter set that minimizes this loss is then selected. In our implementation, we set $H_{\max}=\num{1e4}$, though this upper bound can be adjusted if future sensors exceed this resolution.

\subsubsection{Computation of $\hat{o}_x^{(l)}$, $\hat{\theta}_{\text{off}}^{(l)}$, and $L(\hat{H}^{(l)})$ for a given $\hat{H}^{(l)}$}

For each candidate resolution $\hat{H}^{(l)}$, the computation proceeds in three stages: (1) quantify the discrepancy $\Delta\theta_i$ between the observed azimuth and the ideal angle implied by $\hat{H}^{(l)}$ as a function of $\omega_i$, which under the correct resolution forms a piecewise linear function; (2) extract local linear pieces to obtain a prior slope estimate; and (3) refine the offsets by compensating for periodic discontinuities and regressing a continuous linear model whose residuals define the normalized loss $L(\hat{H}^{(l)})$.

For a candidate resolution $\hat{H}^{(l)}$, the ideal azimuths are restricted to the discrete set
\[
\left\{ \tfrac{2\pi}{\hat{H}^{(l)}} \cdot k \;\middle|\; k \in \{0,1,\dots,\hat{H}^{(l)}-1\} \right\}.
\]
Hence each observed angle $\theta_i$ can be associated with the nearest element of this set via an integer index $k_i$, and the discrepancy is defined as
\begin{equation} \label{eq:delta_theta}
\Delta \theta_i = \theta_i - \frac{k_i}{\hat{H}^{(l)}} \cdot 2\pi, 
\quad k_i = \left\lfloor \frac{\theta_i}{2\pi} \cdot \hat{H}^{(l)} + 0.5 \right\rfloor.
\end{equation}
Substituting $\theta_i$ from Equation~\eqref{eq:theta_approx} into $k_i$ and simplifying we obtain
\[
k_i = \left\lfloor \frac{h^{(l)}}{H^{(l)}} \cdot \hat{H}^{(l)} + \frac{\theta_{\text{off}}^{(l)} + o_x^{(l)} \cdot \omega_i}{2\pi} \cdot \hat{H}^{(l)} + 0.5 \right\rfloor.
\]
Assuming the candidate resolution $\hat{H}^{(l)}$ is correct ($\hat{H}^{(l)} = H^{(l)}$), we obtain
\[
k_i = h^{(l)} + \left\lfloor \frac{\theta_{\text{off}}^{(l)} + o_x^{(l)} \cdot \omega_i}{2\pi} \cdot \hat{H}^{(l)} + 0.5 \right\rfloor,
\]
where we have used that $h^{(l)} \in \mathbb{Z}$ and is therefore unaffected by the rounding operation. We express the rounded value as an integer correction term
\[
k_i = h^{(l)} + \Delta h_i^{(l)}, \qquad 
\Delta h_i^{(l)} := \left\lfloor \frac{H^{(l)}}{2\pi}\big(\theta_{\text{off}}^{(l)} + o_x^{(l)} \omega_i\big) + 0.5 \right\rfloor \in \mathbb{Z}.
\]
Since the only non-constant factor in $\Delta h_i^{(l)}$ is $\omega_i > 0$, $\Delta h_i^{(l)}$ is an integer-valued, piecewise-constant function of $\omega_i$:
\[
\Delta h_i^{(l)} = f(\omega_i), 
\quad f : \mathbb{R}_{>0} \to \mathbb{Z}\;\;\text{piecewise constant}.
\]
Substituting $k_i$ into Equation~\eqref{eq:delta_theta} gives
\[
\Delta \theta_i = \underbrace{\frac{h^{(l)}}{H^{(l)}} \cdot 2\pi + \theta_{\text{off}}^{(l)} + o_x^{(l)} \cdot \omega_i}_{\theta_i \text{ (from Eq.~\eqref{eq:theta_approx})}} - \frac{h^{(l)} + \Delta h_i^{(l)}}{H^{(l)}} \cdot 2\pi,
\]
which simplifies to
\begin{equation} \label{eq:theta_delta}
\Delta \theta_i = - \Delta h_i^{(l)} \cdot \frac{2\pi}{H^{(l)}} + \theta_{\text{off}}^{(l)} + o_x^{(l)} \cdot \omega_i.
\end{equation}
Thus when $\hat{H}^{(l)} = H^{(l)}$, $\Delta \theta_i$ is a piecewise linear function of $\omega_i$, with slope $o_x^{(l)}$ and periodic intercept of period $T^{(l)} = 2\pi/H^{(l)}$.

We estimate the horizontal offset $o_x^{(l)}$ by exploiting the piecewise linearity of $\Delta \theta_i$. We define two segmentation thresholds, $\tau_\omega$ (inverse-range threshold) and $\tau_{\Delta\theta}$ (angular threshold), to partition the $(\omega_i, \Delta \theta_i)$ space into contiguous segments. Consecutive points are assigned to the same segment if the difference in $\omega_i$ is smaller than $\tau_\omega$ and the difference in $\Delta \theta_i$ is smaller than $\tau_{\Delta\theta}$. This procedure isolates regions corresponding to the individual linear portions of the piecewise function. In our implementation, the inverse-range threshold is empirically chosen as $\tau_\omega = 10^{-2}\,\mathrm{m}^{-1}$. The angular threshold is set to $\tau_{\Delta\theta} = T^{(l)}/4$ rad. Since discontinuities between neighboring branches occur in multiples of this period, using one quarter of the period provides a conservative margin: points crossing a discontinuity are separated, while points on the same local linear branch can still be grouped. This segmentation is used only to obtain an initial slope estimate. The final horizontal offset, azimuthal offset, and resolution are determined by the subsequent periodic fitting step and normalized loss.

Within each segment, we perform a linear regression to estimate a local slope. These slope estimates are then aggregated using a weighted median, where each segment contributes proportionally to its number of points. In this way, larger and more reliable segments exert greater influence on the aggregated estimate $\hat{\hat{o}}_x^{(l)}$. This initial estimate serves as a prior for the subsequent optimization steps. We first define a reference line with zero intercept and slope $\hat{\hat{o}}_x^{(l)}$, and compute the residuals of each point $(\omega_i, \Delta \theta_i)$ with respect to this line:
\[
R_i = \Delta \theta_i - \hat{\hat{o}}_x^{(l)} \cdot \omega_i = - \Delta h_i^{(l)} \cdot \frac{2\pi}{H^{(l)}} + \theta_{\text{off}}^{(l)}.
\]

Because $\Delta h_i^{(l)}$ is integer-valued, it contributes multiples of $2\pi/H^{(l)}$. Hence the residuals are congruent with the azimuthal offset modulo this period:
\[
R_i \equiv \theta_{\text{off}}^{(l)} \pmod{2\pi/H^{(l)}}.
\]
The azimuthal offset $\theta_{\text{off}}^{(l)}$ can therefore be estimated by applying a circular mean to the residuals modulo $2\pi/H^{(l)}$:
\[
\hat{\hat{\theta}}_{\text{off}}^{(l)} = \operatorname{CircMean}\!\left(R_i \bmod \tfrac{2\pi}{H^{(l)}}\right).
\]

At this point we obtain preliminary estimates $\hat{\hat{o}}_x^{(l)}$ and $\hat{\hat{\theta}}_{\text{off}}^{(l)}$. To refine them, we compute centered residuals by subtracting the estimated offset:
\[
R'_i = R_i - \hat{\hat{\theta}}_{\text{off}}^{(l)} \approx - \Delta h_i^{(l)} \cdot \frac{2\pi}{H^{(l)}}.
\]
From these, the correction terms $\Delta h_i^{(l)}$ are estimated as
\[
\hat{\Delta} h_i^{(l)} = - \left\lfloor\frac{H^{(l)}}{2\pi} R'_i + 0.5\right\rfloor.
\]
Assuming a correct estimation, we have $\hat{\Delta} h_i^{(l)} = \Delta h_i^{(l)}$. This equality holds under the assumptions of $\hat{H}^{(l)} = H^{(l)}$ and $\hat{\hat{o}}_x^{(l)}\approx o_x^{(l)}$. We transform the original piecewise-linear $\Delta \theta_i$ into a continuous linear form by compensating for the periodic discontinuities:
\[
\hat{\Delta} \theta_i = \Delta \theta_i + \hat{\Delta} h_i^{(l)} \cdot \frac{2\pi}{H^{(l)}}.
\]
Using the definition of $\Delta \theta_i$ from Equation~\eqref{eq:theta_delta}, this simplifies to
\begin{equation}
\hat{\Delta} \theta_i = \theta_{\text{off}}^{(l)} + o_x^{(l)} \cdot \omega_i.    
\end{equation}
This linear form explicitly shows that $\hat{\Delta} \theta_i$ depends linearly on $\omega_i$, with slope $o_x^{(l)}$ and intercept $\theta_{\text{off}}^{(l)}$. Consequently, performing linear regression on the transformed pairs $(\omega_i, \hat{\Delta} \theta_i)$ yields refined estimates $\hat{o}_x^{(l)}$ and $\hat{\theta}_{\text{off}}^{(l)}$. The loss metric for the fit is quantified by the squared sum of the residuals,
\begin{equation}
L(\hat{H}^{(l)}) = \sum_i (R''_i)^2 \cdot (\hat{H}^{(l)})^2,    
\end{equation}
where
\[
R''_i = \hat{\Delta} \theta_i - \hat{\theta}_{\text{off}}^{(l)} - \hat{o}_x^{(l)} \cdot \omega_i.
\]

Note that the factor $(\hat{H}^{(l)})^2$ normalizes the loss across candidate resolutions. Since the residuals $R''_i$ are bounded by the period $2\pi/H^{(l)}$, their magnitudes decrease with increasing resolution. Without normalization, higher $\hat{H}^{(l)}$ values would be artificially favored.

\subsubsection{Heuristic Fallback}
For scanlines containing insufficient points, the estimation procedure described above may be unreliable due to limited data. To address this, such scanlines are deferred until all others have been processed, after which their parameters are estimated using a heuristic fallback strategy.

This strategy reuses information from previously estimated scanlines. Specifically, we collect the values of $H^{(l)}$ and $o_x^{(l)}$ obtained from other beams, and for each candidate pair $(\hat{H}^{(l)}, \hat{o}_x^{(l)})$ we compute a heuristic loss function $M(\hat{H}^{(l)}, \hat{o}_x^{(l)})$. The parameter set minimizing this loss is selected as the final estimate.

We begin by computing corrected angles by removing the effect of the horizontal offset:
\[
\theta_i^{\text{corrected}} = \theta_i - \theta_{\text{res}}^{(l,h^{(l)})} \approx \theta_i - \hat{o}_x^{(l)} \cdot \omega_i.
\]
Substituting in Equation~\eqref{eq:theta_approx}, if the candidate $\hat{o}_x^{(l)}$ is correct, the corrected angles satisfy
\[
\theta_i^{\text{corrected}} \approx \frac{h^{(l)}}{H^{(l)}} \cdot 2\pi + \theta_{\text{off}}^{(l)},
\]
so the corresponding ideal quantized angles implied by $\hat{H}^{(l)}$ are:
\[
\theta_i^* = \left\lfloor \frac{\theta_i^{\text{corrected}}}{2\pi/\hat{H}^{(l)}} + 0.5 \right\rfloor \cdot \frac{2\pi}{\hat{H}^{(l)}}.
\]
For correct parameters, the corrected angles should align with the ideal quantization grid up to a constant $\theta_{\text{off}}^{(l)}$, i.e.,
\[
\forall i: \quad \theta_i^* \approx \theta_i^{\text{corrected}} - \theta_{\text{off}}^{(l)}.
\]
This observation allows us to estimate the azimuthal offset as the mean deviation between corrected and ideal angles
\[
\hat{\theta}_{\text{off}}^{(l)} = \frac{1}{n}\sum_{i=1}^{n}\left(\theta_i^{\text{corrected}} - \theta_i^*\right).
\]
Finally, the quality of each candidate parameter set is quantified by the mean absolute deviation
\begin{equation}
M(\hat{H}^{(l)}, \hat{o}_x^{(l)}) = \frac{1}{n}\sum_{i=1}^{n}\left|\theta_i^{\text{corrected}} - \theta_i^* - \hat{\theta}_{\text{off}}^{(l)}\right| \cdot \hat{H}^{(l)}.
\end{equation}

As in the main estimation procedure, the multiplication by $\hat{H}^{(l)}$ serves as a normalization factor. Since the residual magnitudes shrink with increasing resolution due to the shorter period $2\pi/H^{(l)}$, this scaling ensures that heuristic loss values remain comparable across candidate resolutions.

The candidate pair $(\hat{H}^{(l)}, \hat{o}_x^{(l)})$ minimizing $M$ is selected as the final estimate. This fallback provides reasonable and consistent parameter estimates even with limited data.

\subsection{Range Image Generation and Point Cloud Reconstruction}
Once all intrinsic parameters have been estimated, ALICE-LRI can generate lossless range images from point clouds and subsequently reconstruct the original point clouds without information loss. This bidirectional transformation constitutes the core objective of our approach.

\subsubsection{Point Cloud to Range Image} \label{sec:point_cloud_to_range_image}
Given estimated intrinsics $\{\varphi^{(l)}, o_y^{(l)}, H^{(l)}, o_x^{(l)}, \theta_{\text{off}}^{(l)}\}_{l=0}^{L-1}$ and a point cloud $\{(x_i,y_i,z_i)\}$, we first compute spherical coordinates $(r_i, \varphi_i, \theta_i)$ as in Equation~\eqref{eq:ideal_spherical}. Subsequently, each point is assigned to the candidate beam $\hat{l}$ whose vertical parameters best explain its elevation angle $\varphi_i$:
\[
\hat{l}_i = \arg\min_{l} \left|\varphi_i - \varphi^{(l)} - \arcsin\left(\frac{o_y^{(l)}}{r_i}\right)\right|.
\]
The corrected azimuthal angle $\theta_i'$ is computed using the second expression in Equation~\eqref{eq:angle_corrections}:
\[
\theta_i' = \theta_i - \arcsin\left(\frac{o_x^{(\hat{l})}}{r_i \cos(\varphi_i)}\right) - \theta_{\text{off}}^{(\hat{l})}.
\]

The height of the range image is $L$, and the width is $H = \operatorname{LCM}\{H^{(l)}\}$ (least common multiple) to ensure a uniform grid across all beams even when their resolutions differ. The horizontal ($u_i$) and vertical ($v_i$) indices are computed as
\[
\left\{
\begin{aligned}
u_i &= \left\lfloor \frac{\theta_i'}{2\pi} \cdot H + 0.5 \right\rfloor \\
v_i &= L - \hat{l}_i - 1 \quad \text{(see Eq.~\eqref{eq:beam_ordering})},
\end{aligned}
\right.
\]
where uniqueness of $(u_i,v_i)$ is guaranteed under correct parameters. This uniqueness condition prevents many-to-one conflicts, so no overwrite or nearest-range tie-breaking rule is required under the recovered sensor-consistent mapping. Finally, each $r_i$ is stored on pixel $(u_i,v_i)$.

\subsubsection{Range Image to Point Cloud}
Given estimated intrinsics $\{\varphi^{(l)}, o_y^{(l)}, o_x^{(l)}, \theta_{\text{off}}^{(l)}\}_{l=0}^{L-1}$ and a range image of size $H \times L$, each non-empty pixel $(u_i,v_i)$ with range $r_i$ can be back-projected to a 3D point. First, the beam index is identified as $\hat{l}_i=L - v_i - 1$ and then the original observed angles are recovered by adding back the angular residuals:
\[
\left\{
\begin{aligned}
\varphi_i &= \varphi^{(\hat{l}_i)} + \arcsin\!\left(\frac{o_y^{(\hat{l}_i)}}{r_i}\right) \\
\theta_i &= \frac{2\pi u_i}{H} + \theta_{\text{off}}^{(\hat{l}_i)} + \arcsin\!\left(\frac{o_x^{(\hat{l}_i)}}{r_i\cos\varphi_i}\right).
\end{aligned}
\right.
\]
Finally,
\[
x_i = r_i\cos(\varphi_i)\cos(\theta_i),\quad
y_i = r_i\cos(\varphi_i)\sin(\theta_i),\quad
z_i = r_i\sin(\varphi_i).
\]
Under the assumption of correctly estimated intrinsics, this reconstruction is lossless up to floating-point precision.

\subsection{Implementation Settings and Reproducibility} \label{sec:implementation_settings}
For reproducibility, Table~\ref{tab:implementation_settings} provides a compact summary of the main numerical settings and thresholds used in the parameter-estimation stage. Data-dependent quantities such as $n^{(l)}$ and $T^{(l)}$ are computed independently for each scanline candidate.

\begin{table}[H]
    \centering
    \footnotesize
    \caption{Main numerical settings and thresholds used in the ALICE-LRI implementation.}
    \begin{tabular}{@{}p{0.27\columnwidth}p{0.39\columnwidth}p{0.21\columnwidth}@{}}
\toprule
Symbol & Value & Reference \\
\midrule
\(O_y\) & \(\SI{0.5}{\meter}\) & Sec.~\ref{sec:hough_transform} \\
\(\Delta\tilde{o}_y\) & \(\SI{1e-3}{\meter}\) & Sec.~\ref{sec:hough_transform} \\
\(\Phi\) & \(\pi/2\) & Sec.~\ref{sec:hough_transform} \\
\(\Delta\tilde{\varphi}'\) & \(\SI{1e-4}{\radian}\) & Sec.~\ref{sec:hough_transform} \\
\(\epsilon_{\min}\) & \(\SI{1e-6}{\meter}\) & Sec.~\ref{sec:error_bounds} \\
\(N_{\mathrm{WLS}}\) & \(10\) iterations & Sec.~\ref{sec:vertical_fitting_wls} \\
\(H_{\max}\) & \(\num{1e4}\) & Sec.~\ref{sec:horizontal_estimation} \\
\(\tau_\omega\) & \(\SI{1e-2}{\per\meter}\) & Sec.~\ref{sec:horizontal_estimation} \\
\(\tau_{\Delta\theta}\) & \(T^{(l)}/4\) & Sec.~\ref{sec:horizontal_estimation} \\
\bottomrule
\end{tabular}

    \label{tab:implementation_settings}
\end{table}

\section{Materials and Methods} \label{sec:materials_methods}
This section details the experimental methodology used to evaluate the proposed approach. We describe the datasets employed, the experimental setup, and the metrics used for quantitative and qualitative assessments.

\subsection{Datasets}
The proposed method was evaluated on two publicly available datasets: KITTI~\cite{geiger2013kitti} and DurLAR~\cite{li2021durlar}. These datasets were selected to represent two distinct sensor configurations with contrasting characteristics, enabling thorough evaluation of the proposed approach. Their main characteristics are provided in Table~\ref{tab:datasets}.

KITTI features point clouds acquired with a Velodyne HDL-64E sensor with 64 scanlines and substantial beam offsets (vertical: \SI{100}{mm} to \SI{210}{mm}, horizontal: \SI{-26}{mm} to \SI{26}{mm}). The dataset presents challenges primarily due to these significant geometric distortions.

DurLAR provides point clouds from an Ouster OS1-128 sensor with 128 scanlines and minimal offsets (vertical: \SI{25}{mm} to \SI{40}{mm}, horizontal: \SI{-1}{mm} to \SI{1}{mm}). Despite the smaller geometric distortions, DurLAR poses distinct challenges due to its higher scanline count and extended vertical field of view ($[\SI{-22.5}{\degree}, \SI{22.5}{\degree}]$). The wider angular coverage increases the likelihood of missing returns at high elevation angles where ground surfaces are absent, resulting in sparse or missing scanlines and varying point densities.

These complementary characteristics allow comprehensive evaluation of the proposed method: KITTI tests robustness against significant sensor non-idealities and geometric distortions, while DurLAR evaluates performance with compact sensors under challenging data sparsity conditions at extreme viewing angles.

\begin{table*}[htpb]
\centering
\caption{Characteristics of the datasets used for evaluation, showing sensor specifications and data distribution.}
\label{tab:datasets}
\begin{tabular}{@{}cccccccc@{}}
\toprule
\multirow{2}{*}{\textbf{Dataset}} & \multirow{2}{*}{\textbf{Sequences}} & \textbf{Point} & \multirow{2}{*}{\textbf{Scanlines}} & \textbf{Horizontal} & \textbf{Vertical} & \textbf{Vertical} & \textbf{Horizontal}\\
 &  & \textbf{Clouds} &  & \textbf{Resolution} & \textbf{Angles} & \textbf{Offsets} & \textbf{Offsets} \\
\midrule
KITTI~\cite{geiger2013kitti}   & $151$                & \num{47885}                  & $64$                 & $4000$                           & $[\SI{-24.9}{\degree}, \SI{2.0}{\degree}]$ & $[\SI{100}{mm}, \SI{210}{mm}]$            & $[\SI{-26}{mm}, \SI{26}{mm}]$  \\
DurLAR~\cite{li2021durlar}  & $5$                 & \num{145916}                  & $128$                & $2048$                           & $[\SI{-22.5}{\degree}, \SI{22.5}{\degree}]$ & $[\SI{25}{mm}, \SI{40}{mm}]$              & $[\SI{-1}{mm}, \SI{1}{mm}]$    \\ \bottomrule
\end{tabular}
\end{table*}

\subsection{Experimental Setup} \label{sec:experimental_setup}
All experiments were conducted on the CESGA Finisterrae III~\cite{cesga_ft3} supercomputing cluster to efficiently process the entire KITTI and DurLAR datasets, which contain \num{193801} point clouds. This allowed us to analyze both datasets in their entirety rather than using only small subsets. For each point cloud, results---including estimated sensor parameters, range images, and evaluation metrics---were stored in local SQLite~\cite{sqlite} databases to ensure reproducibility and facilitate downstream analysis. After processing, these databases were aggregated and analyzed offline on a local workstation using custom scripts.

Runtime performance analysis was conducted separately on a local workstation to provide timing measurements under controlled conditions. The workstation specifications include an Intel Core i7-13700K processor and 64 GB of RAM, running Ubuntu 22.04.5 LTS.

\subsection{Evaluation Metrics} \label{sec:evaluation_metrics}
The evaluation described in Section~\ref{sec:evaluation} assesses whether the inferred projection model preserves the measurement geometry of the original LiDAR scans during range-image conversion. We therefore consider two complementary aspects: (i) the accuracy of the estimated sensor parameters that define this projection model, and (ii) the preservation of measured point samples after 3D--2D--3D conversion. In addition, the application presented in Section~\ref{sec:application} demonstrates the practical benefits through a compression case study. We combine sensor-parameter accuracy metrics, reconstruction errors expressed in physical units, point-preservation metrics, and local geometric error measures, as detailed below.

\paragraph{(i) Sensor Parameter Estimation} 
For parameter estimation accuracy, we use standard classification metrics (overall accuracy, precision, recall, F1-score) and mean absolute error (MAE) for continuous parameters. We report both macro-averaged (mP, mR, mF1) and weighted-averaged (wP, wR, wF1) variants of classification metrics. Macro-averaged metrics compute the unweighted mean across all classes, while weighted-averaged metrics weight each class by its support to reflect class imbalance.

\paragraph{(ii) Point Cloud Reconstruction Quality} 
To evaluate the fidelity of reconstructed point clouds, we employ geometric similarity metrics that compare points in the original set $P$ with their counterparts in the reconstruction $\hat{P}$.

\begin{itemize}
    \item \textbf{Chamfer Distance (CD)}: The Chamfer Distance quantifies point cloud similarity by measuring bidirectional nearest-neighbour distances. Despite its widespread use in 3D vision, implementations vary considerably in the literature.
    
    Some researchers utilize the non-squared Euclidean distance~\cite{wu2021density}, while others employ squared distances to penalize outliers more severely~\cite{heo2022flicr}. Additionally, the directional terms may be either summed~\cite{lin2023hyperbolic} or averaged, as in the Point Cloud Utils library~\cite{point-cloud-utils}.
    
    We adopt the non-squared, averaged Chamfer Distance formulation
    \[
    \text{CD}(P, \hat{P}) = \frac{\text{MAE}(P, \hat{P}) + \text{MAE}(\hat{P}, P) }{2},
    \]
    where
    \[
    \text{MAE}(A, B) = \frac{1}{|A|} \sum_{a \in A} \min_{b \in B} \|a - b\|_2.
    \]

    This approach provides an interpretable metric that directly expresses reconstruction error in physical units (meters), representing the average displacement between corresponding points.

    \item \textbf{Peak Signal-to-Noise Ratio (PSNR)}: Adapted from image processing, PSNR quantifies reconstruction quality by comparing maximum signal power to error power:
    \[
    \text{PSNR}(P, \hat{P}) = 10 \cdot \log_{10}\left(\frac{\text{MAX}^2}{\text{MSE}(P, \hat{P})}\right),
    \]
    where
    \[
    \text{MSE}(A, B) = \frac{1}{|A|} \sum_{a \in A} \min_{b \in B} \|a - b\|_2^2,
    \]

    and $\text{MAX}$ represents the maximum representable range value (that is, \SI{120}{m} for KITTI~\cite{geiger2013kitti,velodyne_hdl64e_s2_datasheet} and \SI{170}{m} for DurLAR~\cite{ouster_os1_datasheet}).

    \item \textbf{Sampling Error (SE)}: Point preservation is quantified through sampling error:
    \[
    \text{SE}(P, \hat{P}) = \frac{\left||P| - |\hat{P}|\right|}{|P|}, \quad 0 \leq \text{SE} \leq 1.
    \]
    This metric measures the relative difference in cardinality between the original and the reconstructed point clouds.

    \item \textbf{Symmetric Point-to-Plane Error (SP2P)}: To assess local geometric preservation, we use a point-to-plane error inspired by point-to-plane metrics for point cloud compression evaluation~\cite{tian2017geometric}. Since the input LiDAR point clouds do not provide surface normals, local normals are estimated by PCA plane fitting over a 12-point nearest-neighbor neighborhood. This choice follows the classical local tangent-plane formulation of Hoppe et al.~\cite{hoppe1992surface} and common point cloud compression evaluation practice when normals are unavailable~\cite{alexiou2019comprehensive}.

    Given a query point cloud $A$ and a reference point cloud $B$, let $b^*(a)$ be the nearest neighbor of $a \in A$ in $B$, and let $n_{b^*(a)}$ be the local normal estimated at that reference point. The directional point-to-plane errors are defined as
    \[
    E_{\mathrm{p2pl}}(A, B) =
    \left\{
    \left|(a - b^*(a)) \cdot n_{b^*(a)}\right|
    \;:\; a \in A
    \right\}.
    \]
    This error is computed in both directions and the resulting values are concatenated into
    \[
    E_{\mathrm{sym}}(P,\hat{P}) =
    E_{\mathrm{p2pl}}(P,\hat{P}) \cup E_{\mathrm{p2pl}}(\hat{P},P).
    \]
    The frame-level SP2P metric is then the mean absolute symmetric point-to-plane error,
    \[
    \mathrm{SP2P}(P,\hat{P}) =
    \frac{1}{|E_{\mathrm{sym}}(P,\hat{P})|}
    \sum_{e \in E_{\mathrm{sym}}(P,\hat{P})} e.
    \]
\end{itemize}

\section{Evaluation} \label{sec:evaluation}

This section evaluates ALICE-LRI on the KITTI and DurLAR datasets. Our analysis combines quantitative metrics with qualitative visualizations to assess estimation accuracy, reconstruction fidelity, and computational performance. To ensure reproducibility, we provide all code and experiment configurations publicly.\footnote{\url{https://github.com/alice-lri/alice-lri-experiments}}

We apply exactly the same implementation and numerical settings summarized in Table~\ref{tab:implementation_settings} across both datasets, without any dataset-specific customization or retuning. Because KITTI and DurLAR involve different sensors, scanline counts, angular layouts, point densities, and coordinate precision, this common configuration provides a practical check against narrow tuning to a single sensor or dataset. We focus on three key aspects of evaluation: (1) parameter estimation accuracy, (2) range image reconstruction quality, and (3) runtime performance for real-time applications. Together, these evaluations assess both the accuracy of the recovered sensor parameters and the fidelity with which measured points are preserved in practical use.

\subsection{Parameter Estimation} \label{sec:parameter_estimation}
First, we evaluate the accuracy of ALICE-LRI in estimating the intrinsic parameters of spinning LiDAR sensors. We assess the estimation of the number of scanlines $L$, the vertical angles $\varphi^{(l)}$, spatial offsets $o^{(l)}$, horizontal resolutions $H^{(l)}$, and azimuthal offsets $\theta_{\text{off}}^{(l)}$.

We run ALICE-LRI on all frames in each dataset and compare the estimated parameters against reference values derived from manufacturer specifications and available sensor calibration data. To ensure consistent evaluation, the reference values for $\varphi^{(l)}$, $o^{(l)}$, $H^{(l)}$, and $\theta_{\text{off}}^{(l)}$ are fixed per dataset according to the known sensor configuration. However, the number of scanlines $L$ may vary between frames because some laser beams occasionally fail to produce returns, requiring frame-specific verification. Since calibration metadata is available for both datasets, we developed dataset-specific verification scripts. These scripts automatically determine the effective scanline count in each frame by checking which of the predefined scanlines receive valid point returns. This procedure ensures an unambiguous, frame-specific reference for $L$, consistent with the documented sensor configuration.

\subsubsection{Scanline Count} \label{sec:scanline_count_evaluation}
We treat the problem of estimating the number of scanlines $L$ as a multi-class classification task, where each sample is a point cloud and each class corresponds to a different number of scanlines.

Table~\ref{tab:scanlines_count_metrics} summarizes the scanline count estimation results on the KITTI and DurLAR datasets. The reported metrics are: number of samples (point clouds), number of incorrect predictions, OA (overall accuracy), mP (macro precision), mR (macro recall), mF1 (macro F1-score), wP (weighted precision), wR (weighted recall), and wF1 (weighted F1-score), all expressed as percentages except for the sample counts. To provide a comprehensive assessment, we partition each dataset into two subsets that highlight performance under different conditions. The ``all'' subset encompasses every point cloud in the dataset, while the ``$n^{(l)} \geq 64$'' subset contains only point clouds with at least 64 points per scanline---a reasonable threshold that maintains most of the dataset while excluding severely degraded scans.

\begin{table*}[htpb]
\centering
\caption{Scanlines count estimation metrics on KITTI and DurLAR datasets. For each subset (all point clouds and those with $n^{(l)} \geq 64$), the number of point clouds (\# Samples), incorrect predictions (\# Incorrect), overall accuracy (OA), macro and weighted precision (mP, wP), recall (mR, wR), and F1-score (mF1, wF1) are reported.}
\begin{tabular}{llrrrrrrrrr}
\toprule
\multirow{2}{*}{\textbf{Dataset}} & \multirow{2}{*}{\textbf{Subset}} & \multicolumn{9}{c}{\textbf{Scanlines Count}} \\
\cmidrule(l){3-11}
 &  & \# Samples & \# Incorrect & OA (\%) & mP (\%) & mR (\%) & mF1 (\%) & wP (\%) & wR (\%) & wF1 (\%) \\
\midrule
\multirow{2}{*}{\textbf{\textbf{KITTI}}} & \textbf{All} & \num{47885} & 0 & 100.00 & 100.00 & 100.00 & 100.00 & 100.00 & 100.00 & 100.00 \\
\textbf{} & \textbf{$n^{(l)} \geq 64$} & \num{47543} & 0 & 100.00 & 100.00 & 100.00 & 100.00 & 100.00 & 100.00 & 100.00 \\
\midrule
\multirow{2}{*}{\textbf{\textbf{DurLAR}}} & \textbf{All} & \num{145916} & 130 & 99.91 & 92.98 & 93.67 & 93.31 & 99.92 & 99.91 & 99.92 \\
\textbf{} & \textbf{$n^{(l)} \geq 64$} & \num{130757} & 0 & 100.00 & 100.00 & 100.00 & 100.00 & 100.00 & 100.00 & 100.00 \\
\bottomrule
\end{tabular}

\label{tab:scanlines_count_metrics}
\end{table*}

The results demonstrate that ALICE-LRI achieves perfect scanline count prediction on KITTI (100\% overall accuracy) and near-perfect results on DurLAR ($99.91\%$ overall accuracy) when considering all point clouds in the datasets. Perfect classification ($100\%$ accuracy) on both datasets is achieved when considering only point clouds with sufficient point density ($\geq$ 64 points per scanline). These results validate the robustness of ALICE-LRI across different sensor configurations and environmental conditions.

The difference between macro metrics (mP, mR, mF1) and their weighted counterparts (wP, wR, wF1) on the DurLAR dataset primarily arises from the relationship between scanline completeness and point density. Point clouds containing all scanlines typically have many points per scanline, making them easier to classify and thus dominant in the weighted metrics. In contrast, point clouds with missing scanlines often contain remaining scanlines with very few points---sometimes only one or two, especially those adjacent to the missing ones. Consequently, these cases are more challenging for the method, and since macro metrics assign equal weight to all classes, they are more affected by such difficult, low-density instances. Therefore, the observed gap between macro and weighted metrics reflects both the underlying data distribution and the expected difficulty of sparse, low-support cases.

\subsubsection{Per-Beam Parameters} \label{sec:per_beam_parameter_evaluation}
We now evaluate the accuracy of the estimated per-beam parameters. For each detected scanline, the estimated horizontal resolutions $\hat{H}^{(l)}$, vertical angles $\hat{\varphi}^{(l)}$, spatial offsets $\hat{o}^{(l)}$, and azimuthal offsets $\hat{\theta}_{\text{off}}^{(l)}$ are compared against the corresponding reference values. Horizontal resolution estimation is treated as a binary classification problem (considered correct if it matches the reference value, and incorrect otherwise) whereas the remaining parameters are evaluated as continuous quantities.

Table~\ref{tab:resolution_metrics} presents the horizontal resolution estimation results across both datasets. The method achieves near-perfect accuracy of 99.99\% on both KITTI (84 incorrect) and DurLAR (237 incorrect) when considering all point clouds in the datasets. When focusing on point clouds with sufficient point density ($n^{(l)} \geq 64$), performance improves further, reaching 100\% accuracy on DurLAR (0 incorrect) and 99.99\% accuracy on KITTI (4 incorrect). These results confirm the robustness of our horizontal resolution estimation approach across different sensor configurations and data quality conditions.

\begin{table}[htpb]
\centering
\caption{Horizontal resolution estimation accuracy on KITTI and DurLAR. We report the total number of scanlines (\# Samples), incorrect predictions (\# Incorrect), and overall accuracy (OA) for all point clouds and for those with $n^{(l)} \geq 64$.}
\begin{tabular}{llrrr}
\toprule
\multirow{2}{*}{\textbf{Dataset}} & \multirow{2}{*}{\textbf{Subset}} & \multicolumn{3}{c}{\textbf{Horizontal Resolution}} \\
\cmidrule(l){3-5}
 &  & \# Samples & \# Incorrect & OA (\%) \\
\midrule
\multirow{2}{*}{\textbf{\textbf{KITTI}}} & \textbf{All} & \num{3064085} & 84 & 99.99 \\
\textbf{} & \textbf{$n^{(l)} \geq 64$} & \num{3042253} & 4 & 99.99 \\
\midrule
\multirow{2}{*}{\textbf{\textbf{DurLAR}}} & \textbf{All} & \num{18637656} & 237 & 99.99 \\
\textbf{} & \textbf{$n^{(l)} \geq 64$} & \num{16736809} & 0 & 100.00 \\
\bottomrule
\end{tabular}

\label{tab:resolution_metrics}
\end{table}

Table~\ref{tab:per_beam_metrics} summarizes the estimation errors for the remaining per-beam parameters using mean absolute error (MAE) and maximum error (MAX) metrics. The results demonstrate high accuracy, particularly on point clouds with adequate point density ($n^{(l)} \geq 64$). For these well-populated scanlines, vertical angle estimation achieves sub-degree accuracy, with MAE values of $4.12 \cdot 10^{-4}$° on KITTI and $8 \cdot 10^{-6}$° on DurLAR, and maximum errors below $0.050$° and $0.013$° respectively. Spatial offset estimation shows remarkable sub-millimeter precision, with MAE values of $5.7 \cdot 10^{-2}$ mm (vertical) and $3.9 \cdot 10^{-2}$ mm (horizontal) on KITTI, and $7.8 \cdot 10^{-4}$ mm and $2.8 \cdot 10^{-3}$ mm on DurLAR. The corresponding maximum errors are well-contained at $4.0$ mm and $19.8$ mm for KITTI, and $0.10$ mm and $0.01$ mm for DurLAR. The high maximum horizontal offset error of $19.8$ mm on KITTI stems from cases with incorrect resolution estimation, as resolution and horizontal offset estimation are interdependent; it reduces to $6.3$ mm when considering only correct resolution estimates. Azimuthal offset estimation is also highly accurate, with MAE values of $8.7 \cdot 10^{-4}$° on KITTI and $2.8 \cdot 10^{-5}$° on DurLAR, and maximum errors of $0.082$° and $1.2 \cdot 10^{-4}$° respectively.

\begin{table*}[htpb]
\centering
\caption{Per-beam parameter estimation errors (MAE and MAX) for vertical angles, vertical and horizontal offsets, and azimuthal offsets on KITTI and DurLAR. Results are shown for all point clouds and for those with $n^{(l)} \geq 64$.}
\label{tab:per_beam_metrics}
\begin{tabular}{llrrrrrrrr}
\toprule
\multirow{2}{*}{\textbf{Dataset}} & \multirow{2}{*}{\textbf{Subset}} & \multicolumn{2}{c}{\textbf{Vert. Angle $\varphi^{(l)}$ (deg)}} & \multicolumn{2}{c}{\textbf{Vert. Offset $o_y^{(l)}$ (mm)}} & \multicolumn{2}{c}{\textbf{Horiz. Offset $o_x^{(l)}$ (mm)}} & \multicolumn{2}{c}{\textbf{Azim. Offset $\theta_{\text{off}}^{(l)}$ (deg)}} \\
\cmidrule(l){3-4} \cmidrule(l){5-6} \cmidrule(l){7-8} \cmidrule(l){9-10}
 &  & MAX & MAE & MAX & MAE & MAX & MAE & MAX & MAE \\
\midrule
\multirow{2}{*}{\textbf{\textbf{KITTI}}} & \textbf{All} & 0.123479 & 0.000413 & 137.338350 & 0.057363 & 202.204798 & 0.040357 & 0.081500 & 0.000872 \\
\textbf{} & \textbf{$n^{(l)} \geq 64$} & 0.049864 & 0.000412 & 4.006125 & 0.056994 & 19.806281 & 0.038512 & 0.081500 & 0.000870 \\
\midrule
\multirow{2}{*}{\textbf{\textbf{DurLAR}}} & \textbf{All} & 3.490045 & 0.000033 & 323.204766 & 0.005478 & 17.107649 & 0.003304 & 0.174496 & 0.000032 \\
\textbf{} & \textbf{$n^{(l)} \geq 64$} & 0.012686 & 0.000008 & 0.102721 & 0.000780 & 0.010094 & 0.002778 & 0.000117 & 0.000028 \\
\bottomrule
\end{tabular}

\end{table*}

When considering all point clouds, MAE values remain low, indicating that the higher maximum errors are caused by a small number of outliers in sparse scans rather than systematic failure. The primary sources of error are insufficient point density, which degrades statistical estimation, and incorrect point-to-scanline assignments in severely degraded scans. Nevertheless, the method demonstrates robust performance on well-populated scanlines, confirming its effectiveness for practical use.

\subsubsection{Ablation Study} \label{sec:ablation_study}
ALICE-LRI incorporates several algorithmic components that vary in mathematical rigor: while core elements like the Hough Transform and weighted least squares fitting have strong theoretical foundations, other components such as the heuristic fallbacks rely on empirical assumptions about LiDAR sensor design. To assess the contribution of each component, we conduct a comprehensive ablation study over the entire KITTI and DurLAR datasets by selectively disabling individual algorithm features. This analysis is crucial for determining whether performance gains arise from principled mathematical modeling or from heuristics that may implicitly encode sensor-specific knowledge.

We focus our ablation analysis on scanline count estimation and horizontal resolution accuracy as they represent the most critical parameters for the internal consistency of the algorithm and parameter interdependencies. Scanline count errors cascade through the entire parameter estimation pipeline, leading to incorrect point-to-scanline assignments that compromise subsequent parameter fitting stages. Similarly, horizontal resolution accuracy is fundamental for proper horizontal and azimuthal offsets estimation, as these parameters are intrinsically coupled in the fitting process---incorrect resolution estimates prevent accurate spatial offset determination. These parameters also exhibit the clearest binary success/failure characteristics, making them ideal metrics for assessing component contributions. While other parameters such as vertical angles and spatial offsets are equally important for reconstruction quality, their continuous nature makes ablation analysis more complex and less interpretable in terms of algorithmic component impact.

We evaluate four key components: (1) \textit{Hough Continuity}, which implements the vote-for-discontinuities strategy to improve robustness to steep gradients in the parameter space; (2) \textit{Conflict Resolution}, which enforces global consistency through backtracking and prevents inconsistent scanline assignments; (3) \textit{Vertical Heuristics}, which provide fallback parameter estimation when weighted least squares fitting fails due to insufficient data; and (4) \textit{Horizontal Heuristics}, which estimate horizontal parameters for sparse scanlines using values from other beams. Note that, while the first two components enhance the core mathematical framework, the latter two represent domain-specific heuristics that exploit typical LiDAR sensor configurations.

When \textit{Hough Continuity} is disabled, the accumulator uses a discontinuous voting scheme as shown in Figure~\ref{fig:discontinuities_regular}. Disabling \textit{Conflict Resolution} reduces the algorithm to a greedy approach where scanlines are immediately accepted upon successful fitting unless they contain already assigned points, in which case they are permanently discarded. When \textit{Vertical Heuristics} are disabled, scanlines that fail weighted least squares estimation due to insufficient data are entirely discarded. Finally, disabling \textit{Horizontal Heuristics} enforces the standard fitting procedure across all scanlines irrespective of point density. Scanlines that fail to converge receive invalid parameter values. The computational overhead of these individual components is negligible compared to the total parameter estimation time, allowing us to focus solely on their algorithmic contributions without performance considerations.

Table~\ref{tab:ablation_combined_metrics} summarizes the features enabled in each experiment configuration (E0--E7), along with their impact on scanline count estimation and resolution accuracy across both datasets. Configuration E0 represents the complete method with all components active, while subsequent configurations (E1-E7) systematically disable specific features to isolate their individual contributions. Values not in parentheses represent all point clouds in the dataset, while values in parentheses represent point clouds with $n^{(l)} \geq 64$ points per scanline. This experimental design allows us to distinguish between improvements achieved through rigorous mathematical methods versus those dependent on heuristic assumptions, thereby validating the fundamental soundness of our approach.

\begin{table*}[htpb]
    \centering
    \caption{Ablation study results showing the impact of different algorithm components on estimation accuracy. For each configuration, we report the number of incorrect scanline counts and resolution errors across KITTI and DurLAR datasets. Values not in parentheses represent all point clouds, while values in parentheses represent point clouds with $n^{(l)} \geq 64$. The full method (E0) includes all components, while subsequent configurations systematically disable specific features.}  \label{tab:ablation_combined_metrics}
    \begin{tabular}{lccccrrrr}
\toprule
 & \multicolumn{4}{c}{\textbf{Algorithm Components}} & \multicolumn{2}{c}{\textbf{KITTI}} & \multicolumn{2}{c}{\textbf{DurLAR}} \\
\cmidrule(l){2-5} \cmidrule(l){6-7} \cmidrule(l){8-9}
 & \multicolumn{1}{c}{Hough} & \multicolumn{1}{c}{Conflict} & \multicolumn{1}{c}{Vertical} & \multicolumn{1}{c}{Horizontal} 
 & \multicolumn{1}{c}{\# Incorrect} & \multicolumn{1}{c}{\# Incorrect} 
 & \multicolumn{1}{c}{\# Incorrect} & \multicolumn{1}{c}{\# Incorrect} \\
 & \multicolumn{1}{c}{Continuity} & \multicolumn{1}{c}{Resolution} & \multicolumn{1}{c}{Heuristics} & \multicolumn{1}{c}{Heuristics}
 & \multicolumn{1}{c}{Scanlines Count} & \multicolumn{1}{c}{Resolutions}
 & \multicolumn{1}{c}{Scanlines Count} & \multicolumn{1}{c}{Resolutions}
 \\
\midrule
\textbf{E0} & \checkmark & \checkmark & \checkmark & \checkmark & 0 (0) & 84 (4) & 130 (0) & 237 (0) \\
\textbf{E1} &  & \checkmark & \checkmark & \checkmark & 0 (0) & 84 (4) & 145 (0) & 260 (0) \\
\textbf{E2} & \checkmark &  & \checkmark & \checkmark & 2 (0) & 86 (4) & 2110 (24) & 3263 (154) \\
\textbf{E3} & \checkmark & \checkmark &  & \checkmark & 136 (3) & 414 (7) & 4137 (0) & 13741 (0) \\
\textbf{E4} & \checkmark & \checkmark & \checkmark &  & 0 (0) & 299 (4) & 130 (0) & 12807 (0) \\
\textbf{E5} & \checkmark & \checkmark &  &  & 136 (3) & 455 (7) & 4137 (0) & 14204 (0) \\
\textbf{E6} &  &  & \checkmark & \checkmark & 5 (2) & 90 (5) & 2294 (7) & 2734 (3) \\
\textbf{E7} &  &  &  &  & 140 (5) & 461 (9) & 5666 (7) & 17694 (4) \\
\bottomrule
\end{tabular}

\end{table*}

The ablation results show that the full method (E0) achieves the best performance, with errors increasing as key components are disabled. Removing core mathematical elements or heuristics leads to substantial drops in accuracy, especially for sparse data. This effect is more pronounced on DurLAR, which presents greater challenges than KITTI. Crucially, disabling heuristics has a very limited impact on dense point clouds in both datasets: even when both vertical and horizontal heuristics are removed, performance on dense data remains nearly perfect.

On DurLAR, configuration E6 (disabling both Hough continuity and conflict resolution) outperforms E2 (disabling only conflict resolution) on dense data. This counterintuitive result stems from specific consecutive frames where Hough continuity creates false accumulator peaks. While conflict resolution can address these issues when enabled, its absence in E2 causes failures that are mitigated in E6 by also disabling the problematic Hough continuity component.

\subsubsection{Robustness Analysis} \label{sec:robustness_analysis}
The ablation results indicate that sparse scanlines are the main regime where additional safeguards matter. We analyze robustness from two complementary perspectives: how often fallback mechanisms are used across the full datasets, and what kind of sparse evidence can reduce the stability of the parameter estimates.

Table~\ref{tab:fallback_frequency} reports how often fallback mechanisms are triggered by the full method (E0), at both frame and scanline level, using the same all/dense partition as the previous evaluation tables. On KITTI, at least one fallback is used in only 153 of 47,885 frames (0.32\%), corresponding to 469 of 3,064,085 scanlines (0.015\%). On DurLAR, fallback use is more frequent at frame level, with 10,507 of 145,916 frames (7.20\%), but it still affects only 143,378 of 18,637,433 scanlines (0.769\%). In the dense subset ($n^{(l)} \geq 64$), no DurLAR frame and only three KITTI frames require any fallback.

\begin{table*}[htpb]
    \centering
    \caption{Fallback-trigger frequency for the full method (E0). The table reports frames and scanlines for which either the vertical or horizontal fallback is used, considering all frames and the dense subset ($n^{(l)} \geq 64$ for every scanline).}
    \begin{tabular}{llrrrr}
\toprule
\multirow{2}{*}{\textbf{Dataset}} & \multirow{2}{*}{\textbf{Subset}} & \multicolumn{2}{c}{\textbf{Frames}} & \multicolumn{2}{c}{\textbf{Scanlines}} \\
\cmidrule(l){3-4} \cmidrule(l){5-6}
 &  & \# Total & \# With Fallback & \# Total & \# With Fallback \\
\midrule
\multirow{2}{*}{\textbf{\textbf{KITTI}}} & \textbf{All} & \num{47885} & \num{153} (0.320\%) & \num{3064085} & \num{469} (0.0153\%) \\
\textbf{} & \textbf{$n^{(l)} \geq 64$} & \num{47543} & \num{3} (0.0063\%) & \num{3042253} & \num{3} (0.0001\%) \\
\midrule
\multirow{2}{*}{\textbf{\textbf{DurLAR}}} & \textbf{All} & \num{145916} & \num{10507} (7.20\%) & \num{18637433} & \num{143378} (0.769\%) \\
\textbf{} & \textbf{$n^{(l)} \geq 64$} & \num{130757} & \num{0} (0\%) & \num{16736809} & \num{0} (0\%) \\
\bottomrule
\end{tabular}

    \label{tab:fallback_frequency}
\end{table*}

\begin{figure*}[!t]
    \centering
    % Compensate for the composite TikZ picture's invisible left extent.
    \hspace*{-2cm}\makebox[\textwidth][c]{\scalebox{0.88}{% Switch this input to compare the alternative layout:
\newcommand{\VerticalWLSFrameId}{150476}
\newcommand{\VerticalWLSFramePath}{DurLAR_20211209/ouster_points/data/0000019028.bin}
\newcommand{\VerticalWLSTrueScanlines}{128}
\newcommand{\VerticalWLSPredictedScanlines}{127}
\newcommand{\VerticalWLSSparseScanlines}{26}
\newcommand{\VerticalWLSMinPoints}{1}
\newcommand{\VerticalWLSHighlightedScanlines}{123,124,126,127}
\newcommand{\VerticalWLSFallbackScanlines}{123,126}
\newcommand{\VerticalWLSBadWLSScanlines}{124}
\newcommand{\VerticalWLSMissingScanlines}{127}
\newcommand{\VerticalWLSMainXMin}{0}
\newcommand{\VerticalWLSMainXMax}{0.34}
\newcommand{\VerticalWLSZoomXMin}{0.0188907324}
\newcommand{\VerticalWLSZoomXMax}{0.0512607531}
\newcommand{\VerticalWLSZoomYMin}{18.55}
\newcommand{\VerticalWLSZoomYMax}{21.7}

\definecolor{verticalwlsTabBlue}{RGB}{31,119,180}
\definecolor{verticalwlsTabCyan}{RGB}{23,190,207}
\definecolor{verticalwlsTabPink}{RGB}{227,119,194}
\definecolor{verticalwlsTabPurple}{RGB}{148,103,189}
\definecolor{verticalwlsTabBrown}{RGB}{118,70,55}
\definecolor{verticalwlsTabGold}{RGB}{188,189,34}
\definecolor{verticalwlsTabTan}{RGB}{210,178,130}
\definecolor{verticalwlsTabOlive}{RGB}{128,128,64}
\definecolor{verticalwlsTabSlate}{RGB}{130,140,146}
\definecolor{verticalwlsTabCharcoal}{RGB}{54,54,54}

\includegraphics{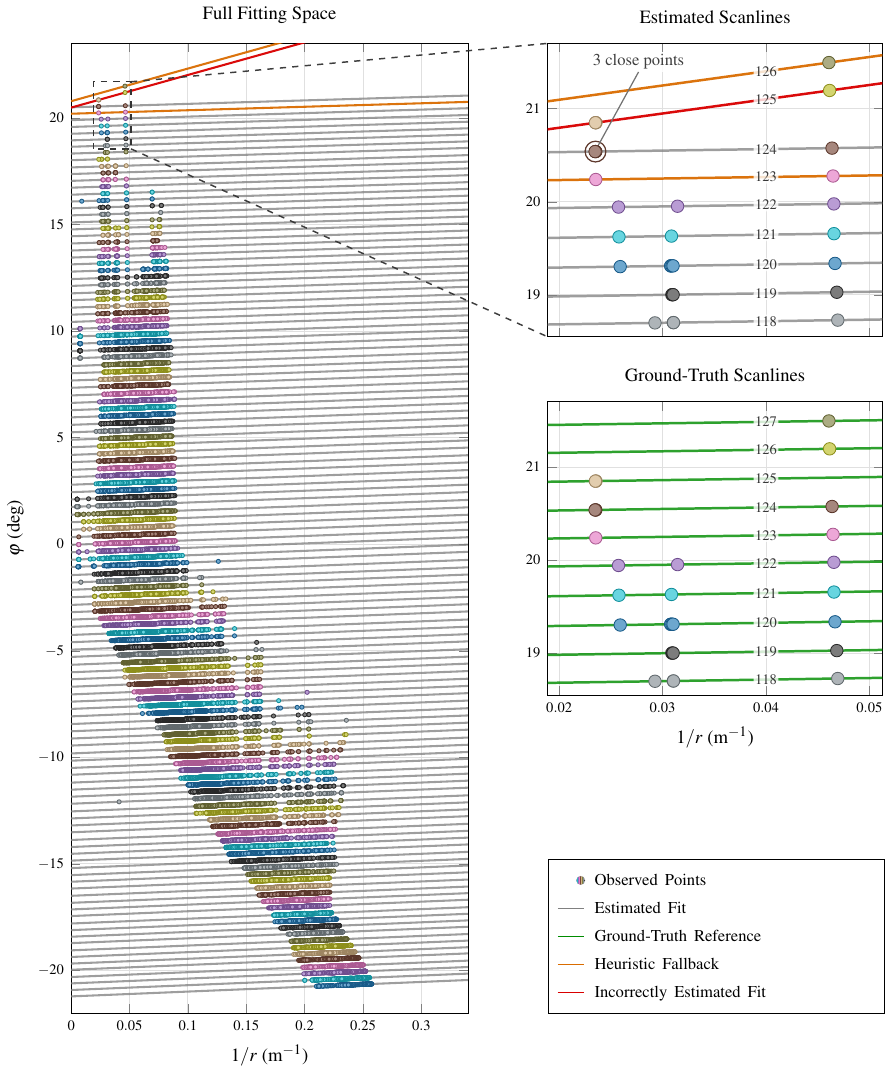}

}}
    \caption{Representative sparse DurLAR frame in the vertical weighted least squares fitting space. Points are plotted as $(1/r,\varphi)$ and colored by scanline using a cyclic categorical palette. The full fitting space shows the global distribution of scanline evidence, while the zoomed panels compare the estimated vertical fits with the corresponding ground-truth scanline references. Gray lines are standard WLS estimates, orange lines are heuristic fallback estimates, the red line is an incorrectly estimated fit, and green lines are ground-truth references. Labels identify scanlines for reference.}
    \label{fig:vertical_wls_fit_space}
\end{figure*}

While these aggregate frequencies show that fallback use is concentrated in sparse scans, they do not, by themselves, explain the underlying failure mode. The dataset-level gap between KITTI and DurLAR is mainly explained by the different sparsity regimes induced by their sensor configurations. As summarized in Table~\ref{tab:datasets}, DurLAR's Ouster OS1-128 has an upper vertical field of view of \SI{22.5}{\degree}, compared with \SI{2.0}{\degree} for KITTI's HDL-64E. Upper DurLAR beams therefore more often point toward sky or other low-return regions in typical outdoor scenes, increasing the probability of sparse or missing scanlines. More generally, estimation becomes unreliable when the available evidence is insufficient to separate neighboring scanlines in the fitting space. Figure~\ref{fig:vertical_wls_fit_space} illustrates this mechanism with a representative sparse DurLAR point cloud transformed into the same linearized WLS space used in Sec.~\ref{sec:vertical_fitting_wls}. After defining $q_i=1/r_i$, each point contributes one observation $(q_i,\varphi_i)$, and points from a scanline should align along the line in Equation~\eqref{eq:linear_model}, whose intercept is the vertical angle and whose slope is the vertical offset.

This transformed point cloud makes the sparsity pattern visible. Lower scanlines contain many observations because those beams point toward the ground and other regions with abundant returns. As elevation increases, the scanlines become progressively sparser, until the upper highlighted region contains too little evidence to separate all beams reliably. The zoomed panels isolate this problematic region.

Scanline 123, shown as an orange fallback line with only two supporting points, is recovered correctly by the sparse-data fallback. Reliable neighboring support scanlines help identify the appropriate returns, and the resulting two points define the line direction algebraically. However, they cannot provide a meaningful WLS uncertainty estimate for a two-parameter fit. Scanline 125, shown in red, is a problematic case. Compared with the green ground-truth references, this fit merges evidence from neighboring sparse scanlines rather than separating them cleanly. The error therefore affects both the estimated vertical offset, which corresponds to the slope in the WLS space, and the scanline count, because two physical scanlines are effectively collapsed into one estimated line. The incorrect estimate also propagates to the fallback estimate for scanline 126: with only one point, this scanline cannot define a line by itself and must be extrapolated from its neighbors.

The ``3 close points'' annotation in scanline 124 illustrates a related issue: clustered observations over a narrow range of $1/r$ provide limited leverage for estimating a stable line. In this particular frame, the ambiguity is resolved by an additional point farther along the $1/r$ axis. If only the three clustered observations were available, however, the vertical fit would again be unreliable. Similar situations can occur whenever the available returns are too few or too concentrated in the fitting space. Thus, the remaining errors should be interpreted as sparse evidence limitations rather than a lack of robustness on normal scans.

Taken together, the ablation study, fallback-frequency analysis, and sparse-frame visualization reveal a clear pattern: the core mathematical framework demonstrates robust performance on well-populated scanlines, while heuristic components serve as essential safeguards for sparse data scenarios. The stark contrast between performance on all point clouds versus dense subsets when no heuristics are used confirms that estimation accuracy fundamentally depends on point density. The aggregate fallback frequencies further show that these safeguards are rarely invoked during normal dense operation, but are important for degraded or incomplete scanlines. When sufficient data is available, the underlying mathematical formulation operates reliably even without domain-specific optimizations, demonstrating the robustness of our approach.

\subsection{Range Image Error Analysis} \label{sec:range_image_error_analysis}
We now turn to the core objective that motivates this work: assessing whether our inferred sensor parameters enable more accurate range image generation compared to standard approaches. While the parameter estimation accuracy demonstrated in the previous subsection validates the technical correctness of our inference algorithm, the true test lies in whether these parameters translate into measurably improved range image quality and lossless reconstruction capability. This evaluation directly addresses the central promise of our approach: enabling lossless range image generation from calibrated point clouds without requiring manufacturer metadata.

For each sequence in the datasets, we first apply our algorithm to the initial point cloud to infer the sensor parameters, then reuse these parameters consistently for all subsequent point clouds in the sequence. This mirrors typical real-world deployment scenarios where practitioners encounter an unknown sensor and must establish its configuration for subsequent lossless range image generation.

Because our evaluation asks whether the range image mapping can be recovered after metadata has been lost, we restrict quantitative baselines to methods that do not use manufacturer metadata, raw packet data, beam identifiers, or preserved acquisition order. Methods that use such information~\cite{dong2021revisiting,triess2020scan} are not directly comparable in this setting: when the intrinsic sensor information required by the projection is available, they can recover a lossless mapping, but they do not address the metadata-free setting targeted by ALICE-LRI. We therefore employ the Projection-By-Elevation-Angle (PBEA) method described by Wu et al.~\cite{wu2021detailed}. PBEA assumes the ideal model described in Section~\ref{sec:ideal_sensor_model} and operates by uniformly sampling the elevation angle space between predefined minimum and maximum values, creating an equidistant grid in the vertical dimension while maintaining uniform azimuthal sampling in the horizontal dimension. This approach requires only the specification of vertical and horizontal range image resolutions along with elevation angle bounds. It is the most widely adopted method in practice due to its simplicity and lack of sensor-specific requirements. Furthermore, Wu et al. demonstrated that PBEA outperforms alternative approaches such as Projection-By-Laser-ID (PBID) when sufficient resolution is available, establishing it as a strong and fair baseline for our comparison.

We evaluate range image generation quality using both quantitative metrics and qualitative visual analysis. The quantitative analysis measures reconstruction accuracy through objective metrics, while the qualitative analysis examines the presence of artifacts and structural preservation to assess practical performance.

\subsubsection{Quantitative Analysis} \label{sec:range_image_quantitative_analysis}
To assess reconstruction quality, we project each point cloud to a range image using our method, then unproject it back to 3D space and compute three key global metrics against the original point cloud: Chamfer Distance (CD) and Peak Signal-to-Noise Ratio (PSNR) for geometric fidelity, and sampling error (SE) representing the fraction of points lost during the projection-unprojection process. We then complement these global reconstruction metrics with the local symmetric point-to-plane error (SP2P).

Table~\ref{tab:range_image_metrics} presents a quantitative comparison between our proposed method and the PBEA baseline across multiple range image resolutions, evaluated on the complete datasets without excluding any frames. Traditional range image generation methods like PBEA require practitioners to manually select appropriate resolutions through trial and error, as the optimal resolution depends on unknown sensor characteristics. We therefore use the native resolution of each sensor (4000$\times$64 for KITTI's HDL-64E and 2048$\times$128 for DurLAR's Ouster OS1-128) as the primary baseline, and additionally report progressively increased PBEA resolutions to test whether oversampling can close the reconstruction gap.

At the native resolution, our results show that ALICE-LRI significantly outperforms PBEA across all metrics. On KITTI at 4000$\times$64, PBEA achieves 0.027 m average Chamfer Distance with 8.69\% sampling error, while ALICE-LRI delivers $3.80 \cdot 10^{-4}$ m CD with 0\% sampling error. On DurLAR at 2048$\times$128, PBEA shows 0.024 m average CD and 3.52\% sampling error compared to ALICE-LRI's $1 \cdot 10^{-6}$ m CD and 0\% sampling error. Crucially, for ALICE-LRI, the maximum sampling error is exactly zero across the entire KITTI and DurLAR datasets, that is, no points are ever lost during projection and unprojection. 

Oversampling the PBEA grid improves reconstruction quality, as shown in Table~\ref{tab:range_image_metrics}, but the performance gap relative to ALICE-LRI persists even with substantial resolution increases. Scaling PBEA resolution by a factor of 32 in each dimension---from 4000$\times$64 to 128000$\times$2048 for KITTI and from 2048$\times$128 to 65536$\times$4096 for DurLAR (representing a 1024-fold increase in total pixels)---reduces the Chamfer Distance to the millimeter/sub-millimeter scale. Nevertheless, this approach still yields non-zero sampling error. This demonstrates that even significant increases in resolution cannot fully address the fundamental limitations of the PBEA approach.

The small residual reconstruction errors of ALICE-LRI reflect the finite coordinate precision of the input point clouds. As discussed in Section~\ref{sec:error_bounds}, KITTI raw coordinates are quantized to the nearest millimeter, whereas DurLAR provides finer coordinate precision. This explains why the residual CD values for ALICE-LRI are slightly higher on KITTI despite zero sampling error.

\begin{table*}[htpb]
    \centering
    \caption{Point cloud reconstruction quality metrics comparing ALICE-LRI against PBEA baseline at native sensor resolutions and progressively increased resolutions. Results are computed over the complete datasets without excluding frames. For each method and resolution, average (AVG) and extreme frame-level (MAX/MIN) values for Chamfer Distance (CD), Peak Signal-to-Noise Ratio (PSNR), and Sampling Error (SE) are reported.}
    \label{tab:range_image_metrics}
    \begin{tabular}{llrrrrrr}
\toprule
\multirow{2}{*}{\textbf{Dataset}} & \multirow{2}{*}{\textbf{Method}} & \multicolumn{2}{c}{\textbf{CD (m)}} & \multicolumn{2}{c}{\textbf{PSNR (dB)}} & \multicolumn{2}{c}{\textbf{SE (\%)}} \\
\cmidrule(l){3-4} \cmidrule(l){5-6} \cmidrule(l){7-8}
 &  & AVG & MAX & AVG & MIN & AVG & MAX \\
\midrule
\multirow{7}{*}{\textbf{KITTI}} & PBEA ($\num{4000} \times \num{64}$) & 0.027419 & 0.051127 & 63.056212 & 45.769613 & 8.689915 & 17.652985 \\
 & PBEA ($\num{8000} \times \num{128}$) & 0.012521 & 0.022010 & 72.296912 & 51.615480 & 0.518697 & 3.034373 \\
 & PBEA ($\num{16000} \times \num{256}$) & 0.006156 & 0.010768 & 79.090223 & 52.142493 & 0.126372 & 0.755860 \\
 & PBEA ($\num{32000} \times \num{512}$) & 0.003055 & 0.005281 & 85.929010 & 52.152183 & 0.034610 & 0.196997 \\
 & PBEA ($\num{64000} \times \num{1024}$) & 0.001520 & 0.002664 & 92.654247 & 61.124456 & 0.009202 & 0.091909 \\
 & PBEA ($\num{128000} \times \num{2048}$) & 0.000758 & 0.001335 & 99.450472 & 63.225820 & 0.002254 & 0.033769 \\
 & \textbf{ALICE-LRI} ($\mathbf{4000} \times \mathbf{64}$) & $\mathbf{0.000380}$ & $\mathbf{0.000423}$ & $\mathbf{109.331016}$ & $\mathbf{108.196745}$ & $\mathbf{0.000000}$ & $\mathbf{0.000000}$ \\
\midrule
\multirow{7}{*}{\textbf{DurLAR}} & PBEA ($\num{2048} \times \num{128}$) & 0.023616 & 0.050328 & 70.359540 & 38.323632 & 3.518374 & 11.450134 \\
 & PBEA ($\num{4096} \times \num{256}$) & 0.010813 & 0.021681 & 79.518543 & 39.413287 & 0.303959 & 1.271056 \\
 & PBEA ($\num{8192} \times \num{512}$) & 0.005354 & 0.010646 & 86.099955 & 48.081959 & 0.083244 & 0.525120 \\
 & PBEA ($\num{16384} \times \num{1024}$) & 0.002657 & 0.005335 & 92.606193 & 49.798801 & 0.013953 & 0.193679 \\
 & PBEA ($\num{32768} \times \num{2048}$) & 0.001324 & 0.002615 & 98.833528 & 50.249024 & 0.003500 & 0.084627 \\
 & PBEA ($\num{65536} \times \num{4096}$) & 0.000662 & 0.001458 & 104.882723 & 50.249114 & 0.001278 & 0.038590 \\
 & \textbf{ALICE-LRI} ($\mathbf{2048} \times \mathbf{128}$) & $\mathbf{0.000001}$ & $\mathbf{0.000006}$ & $\mathbf{157.268976}$ & $\mathbf{140.288547}$ & $\mathbf{0.000000}$ & $\mathbf{0.000000}$ \\
\bottomrule
\end{tabular}

\end{table*}

\begin{figure*}[htpb]
\centering
\begin{subfigure}[t]{0.48\textwidth}
\centering
\resizebox{0.86\linewidth}{!}{\includegraphics{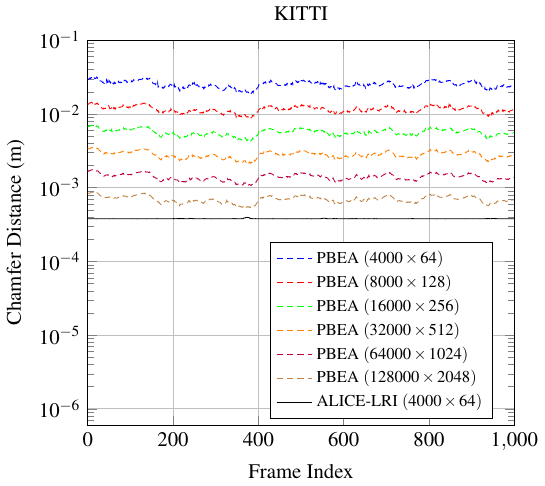}
}
\caption{Chamfer Distance performance across the first 1000 point clouds from the \texttt{2011\_09\_30\_drive\_0018} sequence of the KITTI dataset.}
\label{fig:cd_by_frame_kitti}
\end{subfigure}
\hfill
\begin{subfigure}[t]{0.48\textwidth}
\centering
\resizebox{0.86\linewidth}{!}{\includegraphics{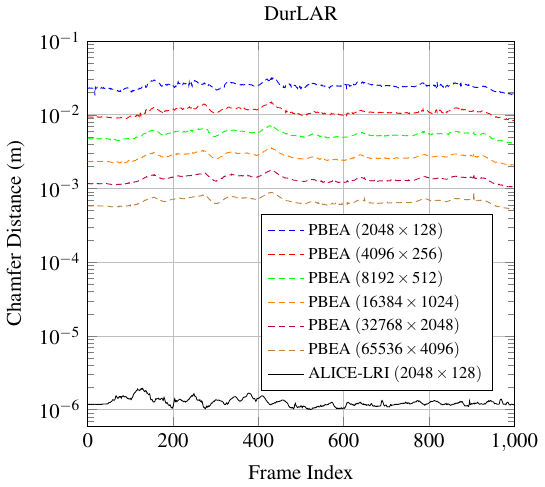}
}
\caption{Chamfer Distance performance across the first 1000 point clouds from the \texttt{DurLAR\_20211209} sequence.}
\label{fig:cd_by_frame_durlar}
\end{subfigure}
\caption{Chamfer Distance of ALICE-LRI and PBEA across different range image resolutions. ALICE-LRI (solid lines) consistently achieves lower reconstruction error compared to PBEA (dashed lines). Results show frame-by-frame reconstruction quality, where lower values indicate better geometric fidelity.}
\label{fig:cd_by_frame}
\end{figure*}

Figure~\ref{fig:cd_by_frame} complements the quantitative results presented in Table~\ref{tab:range_image_metrics} by providing frame-by-frame analysis of reconstruction quality across representative sequences. The plots reveal the consistent performance advantage of ALICE-LRI over the PBEA baseline, with ALICE-LRI (solid lines) maintaining sub-millimeter Chamfer Distance values across all frames. In contrast, the PBEA baseline (dashed lines) exhibits significantly higher reconstruction errors, even at increased resolutions. This frame-level analysis confirms that ALICE-LRI's superior performance is not driven by outliers but represents consistent improvement across diverse scanning conditions and scene contents.

Table~\ref{tab:local_geometry_metrics} complements the global metrics and frame-wise CD inspection by reporting SP2P for ALICE-LRI and PBEA under the same evaluation setting, without excluding frames. The local geometric results are consistent with the global reconstruction analysis: increasing PBEA resolution progressively reduces average SP2P, but even the highest tested resolution remains above ALICE-LRI on both datasets. ALICE-LRI keeps average SP2P close to numerical precision on DurLAR and within the sub-millimeter range on KITTI, whereas native-resolution PBEA produces centimeter-level average local deviations. This indicates that the improvement is not limited to global point-set distances, but also preserves local surface geometry.

\begin{table*}[htpb]
    \centering
    \caption{Symmetric point-to-plane error (SP2P) for ALICE-LRI and PBEA across the evaluated range image resolutions on KITTI and DurLAR. The metric is computed using local PCA normals estimated from a 12-point nearest-neighbor neighborhood. AVG and MAX are computed over frame-level mean absolute SP2P values without excluding frames. Lower values indicate better local geometric preservation.}
    \label{tab:local_geometry_metrics}
    \begin{tabular}{llrr}
\toprule
\multirow{2}{*}{\textbf{Dataset}} & \multirow{2}{*}{\textbf{Method}} & \multicolumn{2}{c}{\textbf{SP2P (m)}} \\
\cmidrule(l){3-4}
 &  & AVG & MAX \\
\midrule
\multirow{7}{*}{\textbf{KITTI}} & PBEA ($\num{4000} \times \num{64}$) & 0.015650 & 0.033118 \\
 & PBEA ($\num{8000} \times \num{128}$) & 0.007421 & 0.016817 \\
 & PBEA ($\num{16000} \times \num{256}$) & 0.003654 & 0.008142 \\
 & PBEA ($\num{32000} \times \num{512}$) & 0.001824 & 0.004040 \\
 & PBEA ($\num{64000} \times \num{1024}$) & 0.000910 & 0.002041 \\
 & PBEA ($\num{128000} \times \num{2048}$) & 0.000454 & 0.001027 \\
 & \textbf{ALICE-LRI} ($\mathbf{4000} \times \mathbf{64}$) & $\mathbf{0.000201}$ & $\mathbf{0.000233}$ \\
\midrule
\multirow{7}{*}{\textbf{DurLAR}} & PBEA ($\num{2048} \times \num{128}$) & 0.010999 & 0.020164 \\
 & PBEA ($\num{4096} \times \num{256}$) & 0.004976 & 0.009076 \\
 & PBEA ($\num{8192} \times \num{512}$) & 0.002479 & 0.004154 \\
 & PBEA ($\num{16384} \times \num{1024}$) & 0.001232 & 0.002076 \\
 & PBEA ($\num{32768} \times \num{2048}$) & 0.000615 & 0.001323 \\
 & PBEA ($\num{65536} \times \num{4096}$) & 0.000307 & 0.000640 \\
 & \textbf{ALICE-LRI} ($\mathbf{2048} \times \mathbf{128}$) & $\mathbf{0.000001}$ & $\mathbf{0.000003}$ \\
\bottomrule
\end{tabular}

\end{table*}

For context, the evaluated sensors report centimeter-level range accuracy or precision: the Velodyne HDL-64E is specified with approximately \SI{2}{cm} distance accuracy~\cite{geiger2013kitti,velodyne_hdl64e_s2_datasheet}, while the Ouster OS1 reports range precision between \SI{0.5}{cm} and \SI{3}{cm} and range accuracy of \SI{2.5}{cm} for Lambertian targets~\cite{ouster_os1_datasheet}. Across the evaluated frames, ALICE-LRI's maximum frame-level CD values are \SI{0.423}{mm} on KITTI and \SI{0.006}{mm} on DurLAR, while the corresponding maximum frame-level SP2P values are \SI{0.233}{mm} and \SI{0.003}{mm}.

Overall, the quantitative analysis confirms that ALICE-LRI achieves lossless range-image conversion with high geometric fidelity. No points are lost in any frame, while CD and SP2P remain well below the relevant manufacturer-reported range-performance specifications of the evaluated sensors. Together, these results demonstrate high-fidelity preservation of the geometric measurement information encoded in the acquired point samples.

\subsubsection{Qualitative Analysis} \label{sec:range_image_qualitative_analysis}
\begin{figure*}[!t]
    \centering
    \begin{subfigure}[t]{0.32\textwidth}
        \centering
        \resizebox{0.88\linewidth}{!}{%
        \includegraphics{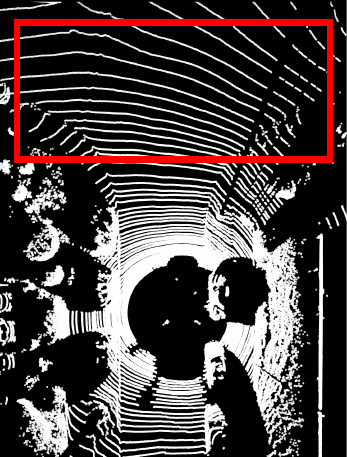}
        }
        \caption{Original point cloud from KITTI dataset showing detailed geometric structure.}
    \end{subfigure}
    \hfill
    \begin{subfigure}[t]{0.32\textwidth}
        \centering
        \resizebox{0.88\linewidth}{!}{%
        \includegraphics{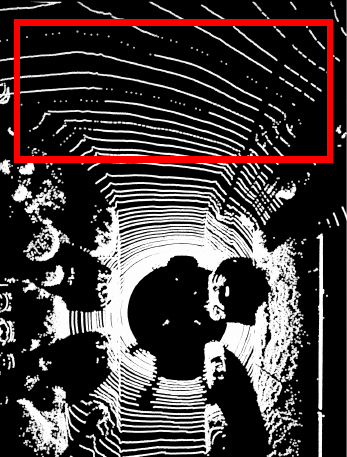}
        }
        \caption{PBEA baseline reconstruction exhibiting geometric distortions and point misalignment (highlighted region).}
    \end{subfigure}
    \hfill
    \begin{subfigure}[t]{0.32\textwidth}
        \centering
        \resizebox{0.88\linewidth}{!}{%
        \includegraphics{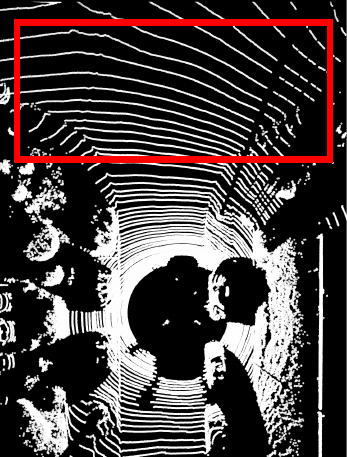}
        }
        \caption{ALICE-LRI reconstruction demonstrating accurate geometric preservation and structural fidelity.}
    \end{subfigure}
    \caption{Qualitative comparison of 3D point cloud reconstruction quality. The highlighted regions (red boxes) illustrate the reconstruction fidelity achieved by ALICE-LRI compared to the baseline PBEA approach. ALICE-LRI maintains structural integrity while the baseline introduces visible artifacts and geometric distortions.}
    \label{fig:qualitative_analysis}
\end{figure*}

The quantitative metrics presented above provide a comprehensive assessment of reconstruction quality, but visual inspection offers complementary insights into the practical implications of using ALICE-LRI. Figure~\ref{fig:qualitative_analysis} shows the differences between the original point cloud and the reconstructions obtained using both the PBEA baseline and ALICE-LRI. The visualizations reveal that while the baseline method introduces geometric distortions and point cloud artifacts, ALICE-LRI preserves the original geometric structure with high fidelity.

\begin{figure*}[!t]
    \centering
    \begin{subfigure}[t]{\linewidth}
        \centering
        \resizebox{0.88\linewidth}{!}{%
        \includegraphics{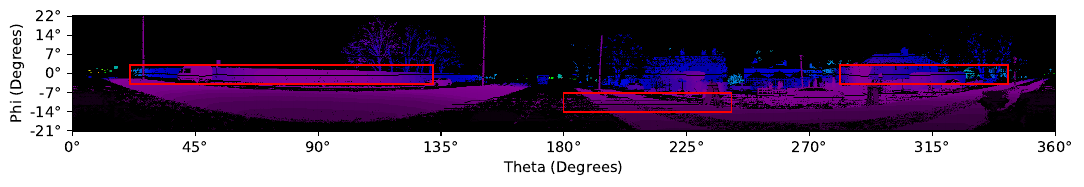}%
        }
        \caption{Range image generated using the PBEA baseline method, showing missing data regions highlighted in red.}
    \end{subfigure}
    \vspace{0.4em}
    \begin{subfigure}[t]{\linewidth}
        \centering
        \resizebox{0.88\linewidth}{!}{%
        \includegraphics{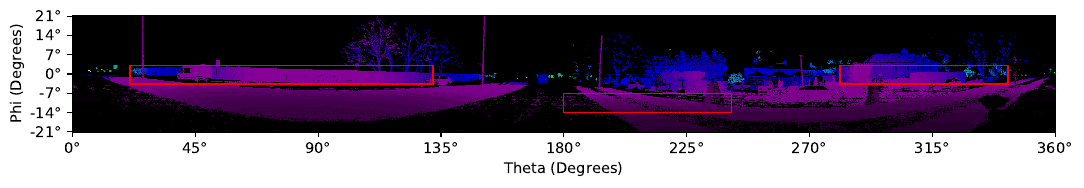}%
        }
        \caption{Range image generated using ALICE-LRI, exhibiting smooth, continuous patterns without artifacts or missing data.}
    \end{subfigure}
    \caption{Comparison of range image quality between baseline and ALICE-LRI methods. The baseline method (top) exhibits visible discontinuities and missing data regions, while ALICE-LRI (bottom) produces clean, artifact-free range images that accurately represent the sensor scanning pattern. The highlighted regions show the improved continuity and data preservation achieved by ALICE-LRI.}
    \label{fig:range_image_comparison}
\end{figure*}

ALICE-LRI not only significantly improves 3D point cloud reconstruction but also enhances the quality of range image representations. Figure~\ref{fig:range_image_comparison} provides a side-by-side comparison of range images generated using both the baseline PBEA and ALICE-LRI. The baseline PBEA method produces range images with missing data regions (highlighted areas). These artifacts arise from the mismatch between the idealized spherical projection model and the actual sensor geometry. In contrast, ALICE-LRI generates range images with smooth, continuous patterns that accurately reflect the underlying sensor scanning behavior. The absence of holes and irregular discontinuities in range images directly correlates with the zero sampling error achieved in our quantitative evaluation, confirming the lossless nature of our approach. This improvement in range image quality has implications beyond reconstruction fidelity: cleaner range images without artifacts can potentially benefit downstream applications such as semantic segmentation, object detection, and scene understanding, though such evaluations lie outside the scope of this work.

\subsection{Runtime Performance} \label{sec:runtime_performance}
Range images often serve as critical intermediate representations in numerous real-time LiDAR processing applications due to their regular grid structure, which enables efficient computation. Since efficiency is critical in such contexts, we now evaluate the runtime performance of ALICE-LRI.

Runtime measurements were obtained by processing representative frames from each dataset on the local workstation described in Section~\ref{sec:experimental_setup}. For KITTI, we used one frame from each of 10 different sequences, while for DurLAR, we used two frames from each of the 5 available sequences. This selection ensures variability across different scene conditions. The implementation operates in single-threaded mode and timing measurements exclude all I/O operations.

Table~\ref{tab:runtime_performance} presents the runtime analysis across both datasets. Parameter estimation requires an average of $31.3$ seconds for KITTI and $41.3$ seconds for DurLAR. This stage is a one-time offline initialization step performed once per sensor, and the recovered parameters are then reused for subsequent frames. In contrast, the real-time processing path consists only of per-frame projection and unprojection. For these per-frame operations, KITTI projection requires an average of $9.3$\,ms per frame, while unprojection completes in $3.8$\,ms. The DurLAR dataset exhibits slightly longer processing times due to its higher scanline count: $20.0$\,ms for projection and $4.8$\,ms for unprojection.

\begin{table}[htpb]
\centering
\caption{Runtime performance analysis showing mean execution times ($\pm$ standard deviation) for parameter estimation, projection, and unprojection phases on the KITTI and DurLAR datasets. Parameter estimation times represent one-time offline estimation costs, while projection and unprojection times indicate per-frame processing requirements for real-time applications.}
\label{tab:runtime_performance}
\begin{tabular}{lrrr}
\toprule
\textbf{Dataset} & \textbf{Estimation (s)} & \textbf{Proj. (ms)} & \textbf{Unproj. (ms)} \\
\midrule
\textbf{KITTI} & $31.3 \pm 0.8$ & $9.3 \pm 0.3$ & $3.8 \pm 0.1$ \\
\textbf{DurLAR} & $41.3 \pm 2.8$ & $20.0 \pm 2.8$ & $4.8 \pm 0.6$ \\
\bottomrule
\end{tabular}

\end{table}

For real-time applications, in the worst case, both the projection and unprojection phases must be executed for every point cloud frame. ALICE-LRI's combined projection and unprojection times are 13.1\,ms for KITTI and 24.8\,ms for DurLAR. Considering that both datasets employ sensors operating at 10~Hz, corresponding to a 100\,ms frame period, these times represent only 13.1\% and 24.8\% of the available budget. This demonstrates that ALICE-LRI comfortably meets real-time constraints while leaving substantial computational headroom for downstream processing.

The final runtime consideration is the one-time parameter estimation stage. Note that this cost is not part of the per-frame real-time path when initialization is performed before operation. If it must instead be amortized over an online session, the additional average cost is $T_{\mathrm{est}}/N$ for a sequence of $N$ frames. At 10~Hz, including both amortized estimation and projection--unprojection, the average per-frame cost falls below the 100\,ms frame budget after approximately 361 frames (36\,s) for KITTI and 550 frames (55\,s) for DurLAR. Thus, the amortized cost remains within the 10~Hz budget for sessions of about one minute or more, and becomes negligible over longer sessions. Nevertheless, ALICE-LRI is not intended for instant operation with a previously unseen sensor and no initialization window.

\section{Application} \label{sec:application}

We now examine the practical implications of ALICE-LRI through a point cloud compression case study. We use the Real-Time Spatio-Temporal Point Cloud Compression (RTST)~\cite{feng2020real} algorithm. RTST meets two useful criteria for this evaluation: (1) it uses range images as an intermediate representation for compression, rather than relying primarily on alternative efficiency mechanisms such as background filtering~\cite{chen2023background,zhao2023decoding}, region-of-interest coding~\cite{sandri2019roi}, or spatial-temporal redundancy coding~\cite{mollah2022efficient}; and (2) it provides open-source code with clearly identifiable projection components that can be replaced with ALICE-LRI. This allows the projection stage to be isolated while keeping all other algorithmic components unchanged.

Compression also provides a controlled setting for validating the main property provided by ALICE-LRI: lossless 3D--2D--3D conversion from calibrated spinning LiDAR point clouds without manufacturer metadata. Projection and unprojection fidelity immediately affect the reconstructed point cloud and can be measured geometrically. By contrast, task-level evaluations in segmentation, detection, or localization would mix the effect of ALICE-LRI's geometry-consistent, lossless range-image representation with task-specific factors such as model architecture, training data, labels, loss functions, and post-processing. In odometry or localization pipelines, such evaluations may also involve motion compensation or extrinsic corrections that are outside the static intrinsic model considered in this work. As a result, isolating the contribution of the range-image mapping would be less direct, whereas compression allows the projection and unprojection stages to be varied while the remaining codec is kept fixed. We therefore use compression as an attribution-friendly representative case study rather than as an exhaustive validation of all downstream tasks.

To ensure a fair comparison, we evaluate the original RTST implementation against a modified version in which only the range image projection and unprojection components are replaced with ALICE-LRI. Although the RTST algorithm is general and applicable to different sensor configurations, the publicly released code by its authors is tailored specifically for the KITTI dataset. To maintain experimental integrity, we chose not to modify the original implementation beyond the projection components. Therefore, this compression evaluation is conducted exclusively on the KITTI dataset.

The RTST algorithm includes an error threshold parameter that controls compression aggressiveness, enabling us to evaluate performance across different compression-quality trade-offs. We compute aggregate metrics over the KITTI dataset, measuring Compression Ratio (CR), Chamfer Distance (CD), Peak Signal-to-Noise Ratio (PSNR), and Sampling Error (SE). The Compression Ratio is defined as:

\[
\text{CR} = \frac{\text{Original Size}}{\text{Compressed Size}}
\]

Table~\ref{tab:compression_results} presents the results of this study. While compression ratios show mixed results, the point cloud quality metrics demonstrate clear advantages for the modified version leveraging ALICE-LRI. At low error thresholds (\numrange{0.001}{0.100}), where reconstruction loss is primarily attributed to projection-unprojection rather than compression artifacts, ALICE-LRI achieves substantial improvements: CD reduces from \SI{0.0293}{\meter} to \SI{0.0020}{\meter} at threshold 0.001, PSNR increases from \SI{62.43}{\decibel} to \SI{93.82}{\decibel}, and SE decreases from \SI{10.09}{\percent} to \SI{0.0009}{\percent}. As error thresholds increase, the performance gap narrows for CD and PSNR since compression artifacts dominate over projection losses. Still, the SE difference remains substantial across all tested configurations, suggesting that most point losses originate from the range image projection-unprojection process rather than the compression algorithm itself.

\begin{table*}[htpb]
    \centering
    \caption{Point cloud compression performance comparison between the original RTST algorithm and the ALICE-LRI modified version. Results show Compression Ratio, Chamfer Distance (CD), Peak Signal-to-Noise Ratio (PSNR), and Sampling Error (SE) across different error thresholds. Bold values indicate superior performance for each metric.}
    \label{tab:compression_results}
    \begin{tabular}{rrrrrrrrr}
\toprule
\multirow{2}{*}{\textbf{Error Threshold}} & \multicolumn{2}{c}{\textbf{Compression Ratio}} & \multicolumn{2}{c}{\textbf{CD (m)}} & \multicolumn{2}{c}{\textbf{PSNR (dB)}} & \multicolumn{2}{c}{\textbf{SE (\%)}} \\
\cmidrule(l){2-3} \cmidrule(l){4-5} \cmidrule(l){6-7} \cmidrule(l){8-9}
 & Original & Modified & Original & Modified & Original & Modified & Original & Modified \\
\midrule
\textbf{0.001} & \textbf{10.7899} & 10.3009 & 0.0293 & \textbf{0.0020} & 62.4274 & \textbf{93.8161} & 10.0882 & \textbf{0.0009} \\
\textbf{0.010} & \textbf{10.4894} & 10.1392 & 0.0293 & \textbf{0.0021} & 62.4270 & \textbf{93.5504} & 10.0881 & \textbf{0.0009} \\
\textbf{0.050} & \textbf{12.2672} & 11.8046 & 0.0315 & \textbf{0.0059} & 62.3669 & \textbf{81.2060} & 10.0877 & \textbf{0.0008} \\
\textbf{0.100} & \textbf{15.0507} & 14.9141 & 0.0362 & \textbf{0.0130} & 62.1628 & \textbf{74.8480} & 10.0873 & \textbf{0.0007} \\
\textbf{0.250} & 19.5483 & \textbf{21.3198} & 0.0474 & \textbf{0.0301} & 61.4164 & \textbf{68.2742} & 10.0869 & \textbf{0.0005} \\
\textbf{0.500} & 23.4855 & \textbf{26.6180} & 0.0627 & \textbf{0.0478} & 60.2123 & \textbf{64.5307} & 10.0867 & \textbf{0.0004} \\
\textbf{0.750} & 26.4736 & \textbf{30.3682} & 0.0760 & \textbf{0.0624} & 59.1101 & \textbf{62.3571} & 10.0866 & \textbf{0.0004} \\
\textbf{1.000} & 28.7996 & \textbf{33.5210} & 0.0878 & \textbf{0.0754} & 58.2241 & \textbf{60.8477} & 10.0865 & \textbf{0.0003} \\
\bottomrule
\end{tabular}

\end{table*}

While the compression ratio results might suggest that ALICE-LRI only improves point cloud quality without enhancing compression efficiency, this interpretation overlooks the practical implications for users. In real-world scenarios, practitioners typically prioritize achieving specific compression ratios while maintaining acceptable point cloud quality rather than optimizing for particular error threshold values. By achieving superior point cloud quality at equivalent compression ratios, ALICE-LRI enables users to select more aggressive compression settings without experiencing significant quality degradation.

\begin{figure}[htpb]
\centering
\includegraphics{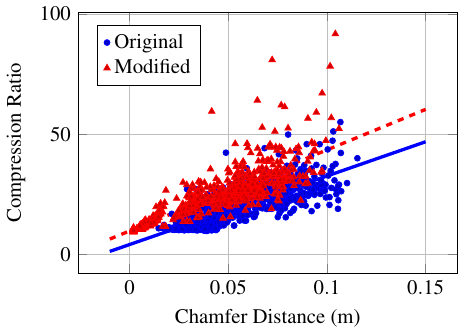}

\caption{Compression Ratio versus Chamfer Distance trade-off analysis. Each point represents one execution of the compression algorithm. Blue circles denote the original method and red triangles denote the ALICE-LRI modified method. The blue solid and red dashed lines show the corresponding linear regression fits, indicating that the ALICE-LRI modified method achieves higher compression ratios on average for equivalent reconstruction quality.}
\label{fig:compression_ratio_vs_cd}
\end{figure}

Figure~\ref{fig:compression_ratio_vs_cd} illustrates this relationship by plotting Compression Ratio against Chamfer Distance for both methods across all evaluated error thresholds. Each point represents one execution of the compression algorithm for either the original or the ALICE-LRI modified method. Linear regression analysis reveals that for a given Chamfer Distance, the modified method achieves higher compression ratios on average compared to the baseline. This result should therefore be interpreted as an improvement in the average quality-compression trade-off, rather than as a guarantee of higher compression ratio under every threshold or operating condition.

These results show that ALICE-LRI enhances range image quality as an intermediate representation and can provide tangible benefits in workflows where projection and reconstruction fidelity directly affect the output. In the compression domain, the main benefit is improved point cloud fidelity and a better quality-compression trade-off, demonstrating the practical value of accurate sensor parameter estimation for real-world LiDAR processing workflows.

\subsection{Runtime Performance Analysis} \label{sec:application_runtime_performance}
Beyond quality improvements, a critical concern for practical deployment is whether ALICE-LRI maintains the computational efficiency required for real-time applications. Section~\ref{sec:runtime_performance} demonstrated substantial computational headroom for ALICE-LRI's projection and unprojection operations. To further validate real-time viability, we analyze the runtime performance impact of integrating ALICE-LRI into the RTST compression workflow.

Table~\ref{tab:rtst_times} presents a detailed comparison of execution times between the original RTST algorithm and our ALICE-LRI modified version across different error thresholds. The results show that ALICE-LRI introduces a modest computational overhead of 2--8\,ms for encoding and 2--3\,ms for decoding operations. Even at the highest error thresholds, where compression times are longest, the total execution times remain well within the computational headroom available for real-time applications, confirming that practitioners can adopt ALICE-LRI without sacrificing temporal performance in time-critical applications.

\begin{table*}[htpb]
    \centering
    \caption{Mean runtime per frame for the original RTST algorithm and the ALICE-LRI modified version across different error thresholds. Execution times are reported in milliseconds. The minimal overhead introduced by ALICE-LRI ensures real-time viability while providing substantial quality improvements.}
    \label{tab:rtst_times}
    \begin{tabular}{rrrrrrr}
\toprule
\multirow{2}{*}{\textbf{Error Threshold}} & \multicolumn{3}{c}{\textbf{Encoding Time (ms)}} & \multicolumn{3}{c}{\textbf{Decoding Time (ms)}} \\
\cmidrule(l){2-4} \cmidrule(l){5-7}
 & Original & Modified & Overhead & Original & Modified & Overhead \\
\midrule
\textbf{0.001} & 20.79 & 23.24 & 2.45 & 3.84 & 6.70 & 2.86 \\
\textbf{0.010} & 19.71 & 26.00 & 6.29 & 4.08 & 6.78 & 2.70 \\
\textbf{0.050} & 22.57 & 28.70 & 6.13 & 4.85 & 7.34 & 2.49 \\
\textbf{0.100} & 30.25 & 37.67 & 7.43 & 4.05 & 7.55 & 3.51 \\
\textbf{0.250} & 43.27 & 49.89 & 6.63 & 4.47 & 7.52 & 3.05 \\
\textbf{0.500} & 56.98 & 65.27 & 8.30 & 4.03 & 7.50 & 3.46 \\
\textbf{0.750} & 67.08 & 72.41 & 5.33 & 4.83 & 7.50 & 2.66 \\
\textbf{1.000} & 75.18 & 80.06 & 4.88 & 4.17 & 7.32 & 3.14 \\
\bottomrule
\end{tabular}

\end{table*}

The minimal runtime overhead observed validates the practical feasibility of ALICE-LRI integration in production systems where both quality and performance are critical requirements.

\section{Conclusion} \label{sec:conclusion}

This work addressed a fundamental limitation in LiDAR-based perception systems: the inability to generate lossless range images from calibrated point clouds without sensor-specific metadata. We presented ALICE-LRI (Automatic LiDAR Intrinsic Calibration Estimation for Lossless Range Images), which automatically estimates the intrinsic calibration parameters governing range-image projection in spinning multi-beam LiDAR sensors and enables truly lossless range image generation when scanlines contain sufficient returns for reliable parameter estimation.

Experimental evaluation on KITTI and DurLAR datasets demonstrated ALICE-LRI's effectiveness across different sensor configurations. Critically, ALICE-LRI achieved zero point loss with reconstruction deviations below the manufacturer-reported range-performance specifications of the evaluated LiDAR sensors during projection and unprojection, while the baseline PBEA method exhibited substantial geometric errors (0.027 m Chamfer Distance on KITTI) and significant sampling losses (8.69\% on KITTI).

The algorithm demonstrated robustness through its iterative approach combining Hough Transform with weighted least squares fitting and backtracking-based conflict resolution. Comprehensive ablation studies revealed that the core mathematical framework performs reliably on well-populated data even without domain-specific heuristics, thereby validating the fundamental soundness of the approach. Beyond accuracy, ALICE-LRI maintained computational efficiency suitable for real-time applications, with projection and unprojection operations completing in milliseconds and achieving substantial computational headroom. In practical applications, the compression case study showed that ALICE-LRI improves point cloud fidelity and the quality-compression trade-off when range images are used as intermediate representations.

ALICE-LRI supports metadata-free use across the evaluated spinning multi-beam LiDAR platforms without dataset-specific retuning. By preserving measured point samples and keeping conversion-induced reconstruction deviations below relevant manufacturer-reported range-performance specifications, the method improves measurement fidelity in range-image-based LiDAR processing and provides a path toward geometry-preserving range image generation. The open-source implementation facilitates adoption in measurement-sensitive applications that require high-precision, lossless LiDAR range-image processing.

The implications extend beyond range image generation to applications requiring accurate geometric modeling of LiDAR data. As autonomous vehicles, robotics, and mapping applications increasingly demand high-fidelity 3D perception, ALICE-LRI provides a foundation for more reliable and accurate LiDAR processing pipelines. The ability to achieve lossless range image generation without manufacturer metadata reduces a practical barrier to deploying LiDAR-based systems across application domains.

\section{Future Work} \label{sec:future_work}
Several promising research directions emerge from our findings. Below, we outline key areas for future investigation that could further advance LiDAR-based perception and geometric modeling.

Our current approach focuses on intrinsic sensor parameters---geometric distortions arising from the physical design and layout of the laser beams. However, in datasets such as KITTI odometry, the ideal spherical projection model is affected not only by intrinsic factors but also by extrinsic corrections applied to compensate for ego-motion during data acquisition. When LiDAR sensors are mounted on moving platforms (e.g., vehicles or robots), point clouds are often motion-corrected to account for sensor displacement and rotation during scanning. These corrections introduce additional geometric transformations that deviate from the static sensor model. Future work could extend our framework to simultaneously estimate intrinsic sensor parameters and extrinsic motion effects. While access to uncorrected raw data or known extrinsic parameters is often available in practice, developing methods that directly account for ego-motion corrections would increase the generality of our approach. This would enable lossless range image generation from a broader class of datasets.

The current formulation is also intended for spinning multi-beam LiDARs with sufficiently populated scanlines. Highly sparse scans, severely incomplete point clouds, or frames with large numbers of missing beams can reduce the reliability of scanline detection and parameter fitting, increasing dependence on heuristic fallbacks. Related deployment factors can exacerbate these limitations: in roadside LiDAR, sensor height, inclination, placement, and occlusions influence which beams receive useful returns, as observed in studies focused on traffic detection~\cite{zhao2022towards}.

Other sensor-geometry boundaries remain outside the present formulation. Highly irregular beam layouts may require adapted parameter bounds or initialization strategies, while non-spinning LiDAR architectures fall outside the rotational sampling model assumed by ALICE-LRI. These cases motivate future extensions to broader sensor geometries and degraded acquisition conditions.

Beyond compression, the impact of lossless range images on downstream perception tasks needs further investigation. Evaluating their effects on semantic segmentation, object detection, and scene understanding could quantify the benefits of eliminating projection artifacts and preserving geometric fidelity. The inferred sensor parameters also open the door to sensor-aware point cloud upsampling and reconstruction. Once sensor geometry is accurately characterized, additional points can be generated in a manner consistent with the physical scanning pattern and measurement characteristics. This capability would be particularly valuable for enhancing sparse automotive LiDAR data and for generating training datasets that accurately reflect sensor-specific properties for machine learning applications.

Finally, even though our current implementation demonstrates real-time viability, it remains single-threaded. Exploiting the inherently parallel nature of per-point projections through multi-threading and GPU acceleration could dramatically reduce execution times, freeing computational resources for downstream tasks and enabling deployment in high-throughput systems.

\section*{CRediT Authorship Contribution Statement}
\textbf{Samuel Soutullo}: Conceptualization, Methodology, Software, Validation, Formal analysis, Investigation, Data curation, Writing -- original draft, Writing -- review \& editing, Visualization. \textbf{Miguel Yermo}: Conceptualization, Methodology, Validation, Writing -- review \& editing. \textbf{David L. Vilariño}: Conceptualization, Validation, Resources, Writing -- review \& editing, Supervision, Project administration, Funding acquisition. \textbf{Óscar G. Lorenzo}: Conceptualization, Validation, Writing -- review \& editing, Supervision, Project administration. \textbf{José C. Cabaleiro}: Conceptualization, Validation, Resources, Writing -- review \& editing, Supervision, Project administration, Funding acquisition. \textbf{Francisco F. Rivera}: Conceptualization, Validation, Resources, Writing -- review \& editing, Supervision, Project administration, Funding acquisition.

\section*{Declaration of Competing Interests}
The authors declare that they have no known competing financial interests or personal relationships that could have appeared to influence the work reported in this paper.

\section*{Data Availability}
The KITTI and DurLAR datasets used in this study are publicly available from their original providers. The ALICE-LRI implementation is publicly available at \url{https://github.com/alice-lri/alice-lri}. The experimental scripts, configurations, derived result databases, and table-generation utilities are publicly available at \url{https://github.com/alice-lri/alice-lri-experiments}.

\section*{Acknowledgments}
The authors acknowledge CESGA (Centro de Supercomputación de Galicia) for providing access to the Finisterrae-III supercomputer, which enabled the computational resources necessary for this research. This research was funded by the Agencia Estatal de Investigación (Spain) (MCIN/AEI/10.13039/501100011033) codes PID2019-104834GB-I00, PID2022-141623NB-I00, PID2024-156264NA-I00, PID2025-174650NB-I00, and PREP2022-000375, the Xunta de Galicia - Consellería de Cultura, Educación, Formación Profesional e Universidades (Centro de investigación de Galicia accreditation 2024-2027 ED431G-2023/04 and Reference Competitive Group accreditation ED431C-2026/04), the European Union (European Regional Development Fund - ERDF/EU).

\section*{Declaration of Generative AI and AI-assisted technologies in the writing process}
During the preparation of this work, the authors used large language model-based AI tools to assist with rephrasing and improving the clarity and structure of the text. After using these tools, the authors reviewed and edited the content as needed and take full responsibility for the content of the publication.

\bibliographystyle{elsarticle-num-names} 
\bibliography{references}

\end{document}